\documentclass[letterpaper, 10 pt, conference]{ieeeconf}  

\IEEEoverridecommandlockouts                              

\usepackage{amsmath} 
\usepackage{amssymb}  
\usepackage{adjustbox}
\usepackage{bm}  
\usepackage{graphics} 
\usepackage{epsfig} 
\usepackage{wrapfig}
\usepackage{subcaption}
\usepackage[font=small,labelfont=small]{caption}
\usepackage{hyperref}
\usepackage{color}
\usepackage[table]{xcolor}
\usepackage[ruled, linesnumbered, vlined]{algorithm2e}
\usepackage{sidecap}
\usepackage{tikz}
\usetikzlibrary{arrows.meta}

\usepackage[per-mode=fraction]{siunitx}
\usepackage{comment}
\usepackage[normalem]{ulem}
\usepackage{xspace}
\usepackage{xcolor}

\DeclareSIUnit\px{px}
\DeclareSIUnit\fps{fps}

\definecolor{OliveGreen}{RGB}{0,200,25}
\newcommand{\red}[1]{\textcolor{red}{#1}}

\newcommand{\darkgreen}[1]{\textcolor{OliveGreen}{#1}}

\newcommand{\armarVI}{\mbox{ARMAR-6}\xspace}

\newif\iffinal

\newcommand{\replaced}[2]{%
	\iffinal%
	#2%
	\else%
	\red{\ifmmode\text{\sout{\ensuremath{#1}}}\else\sout{#1}\fi}\darkgreen{#2}%
	\fi%
}
\newcommand{\removed}[1]{%
	\iffinal%
	\else%
	\red{\ifmmode\text{\sout{\ensuremath{#1}}}\else\sout{#1}\fi}%
	\fi%
}

\graphicspath{{Figures/}}

\title{\LARGE \bf
A Riemannian Take on Human Motion Analysis and Retargeting
}

\author{Holger Klein, No\'emie Jaquier, Andre Meixner, and Tamim Asfour
\thanks{This work was supported by the Carl Zeiss Foundation through the JuBot project.  The authors are with the Institute for Anthropomatics and Robotics, Karlsruhe Institute of Technology, Karlsruhe, Germany. Correspondence to: {\tt\small holgerhugoklein@gmail.com, \{noemie.jaquier, andre.meixner, asfour\}@kit.edu}. }%
}

\newcommand{\trsp}{\mathsf{T}}

\newcommand{\Rho}{\mathrm{P}}

\newcommand{\euclideanspace}{\mathbb{R}}
\newcommand{\manifold}{\mathcal{M}}
\newcommand{\configmanifold}{\mathcal{Q}}
\newcommand{\tangentspace}[1]{\mathcal{T}_{#1}\mathcal{M}}
\newcommand{\configtangentspace}[1]{\mathcal{T}_{#1}\mathcal{Q}}

\newcommand{\innerprod}[3]{\langle #2, #3 \rangle_{#1}}  
\newcommand{\norm}[2]{\| #2\|_{#1}}  
\newcommand{\expmap}[2]{\text{Exp}_{#1}(#2)}  
\newcommand{\logmap}[2]{\text{Log}_{#1}(#2)}  
\newcommand{\prltrsp}[3]{\Rho_{#1 \rightarrow #2}\big(#3\big)}  
\newcommand{\jointposition}{\bm{q}}
\newcommand{\jointvelocity}{\dot{\bm{q}}}
\newcommand{\jointacceleration}{\ddot{\bm{q}}}

\definecolor{darkyellow}{rgb}{0.86, 0.66, 0.}
\definecolor{darkred}{rgb}{0.66, 0.08, 0.23}
\definecolor{darkorange}{rgb}{1., 0.65, 0.0}
\definecolor{skyblue}{rgb}{0., 0.65, 0.9}
\definecolor{steelblue}{rgb}{0.275, 0.501, 0.706}
\definecolor{gray}{rgb}{0.7, 0.7, 0.7}

\DeclareRobustCommand{\blackline}{\raisebox{2pt}{\tikz{\draw[black,solid,line width = 1.5pt](0,0) -- (3mm,0);}}}
\DeclareRobustCommand{\redline}{\raisebox{2pt}{\tikz{\draw[darkred,densely dashdotted,line width = 1.5pt](0,0) -- (3mm,0);}}}
\DeclareRobustCommand{\orangeline}{\raisebox{2pt}{\tikz{\draw[darkorange,dashed,line width = 1.5pt](0,0) -- (3mm,0);}}}
\DeclareRobustCommand{\blueline}{\raisebox{2pt}{\tikz{\draw[steelblue,dashed,line width = 1.5pt](0,0) -- (3mm,0);}}}
\DeclareRobustCommand{\redrectangle}{\tikz{ \filldraw[color=white, fill=darkred!60, thick](0,0) rectangle (.2,.3);}}
\DeclareRobustCommand{\bluerectangle}{\tikz{ \filldraw[color=white, fill=steelblue!60, thick](0,0) rectangle (.2,.3);}}

\begin{document}

\maketitle
\thispagestyle{empty}
\pagestyle{empty}

\begin{abstract}
Dynamic motions of humans and robots are widely driven by posture-dependent nonlinear interactions between their degrees of freedom. However, these dynamical effects remain mostly overlooked when studying the mechanisms of human movement generation. Inspired by recent works, we hypothesize that human motions are planned as \emph{sequences of geodesic synergies}, and thus correspond to coordinated joint movements achieved with piecewise minimum energy. The underlying computational model is built on Riemannian geometry to account for the inertial characteristics of the body. Through the analysis of various human arm motions, we find that our model segments motions into geodesic synergies, and successfully predicts observed arm postures, hand trajectories, as well as their respective velocity profiles.
Moreover, we show that our analysis can further be exploited to transfer arm motions to robots by reproducing individual human synergies as geodesic paths in the robot configuration space. 
\end{abstract}

\section{INTRODUCTION}
\label{sec:Introduction}
Humans exhibit outstanding manipulation and locomotion capabilities while performing a variety of tasks e.g., jumping to reach a highly-situated object, or
climbing stairs holding a cumbersome item. In order to generate such skillful motions, the central nervous system (CNS) is able to efficiently plan optimal trajectories in the task space, while coping with the high redundancy of the configuration space of the human body. In that regard, understanding the underlying mechanisms of human motion generation is key to provide robots with human-like abilities. Indeed, uncovering the CNS internal optimization principles may contribute to the generation of efficient, well-coordinated robot motions, which may additionally be interpretable and predictable by humans.

Various models have been developed in the literature to explain the underlying principles of human motion generation. They are commonly based on the optimization of underlying features (e.g., jerk~\cite{Flash85:MinimumJerk}, acceleration~\cite{Pham07:MinimumAcceleration}, torque change~\cite{Uno89:MinimumTorqueChange}, or combinations of criteria~\cite{Berret2011:PointToManifoldAnalysis}), eventually incorporated in a feedback control loop, to achieve the movement goal (see e.g.,~\cite{Todorov04:NatureOptimalityPrinciplesSensorimotorControl} for a review). However, these models do not account for the dynamical effects inherent to multi-linked mechanical systems such as humans\footnote{From a biomechanical perspective, human bodies are multi-linked mechanical systems with multiple joints.} and robots.
In fact, dynamic movements of such systems are driven by strong posture-dependent non-linearities and couplings among joints. These nonlinear interactions correspond to the inertial and Coriolis terms in the well-known dynamic equation of motion. In that sense, human and robot motions cannot suitably be treated using classical linear, i.e., Euclidean, methods.

The use of geometric methods has been investigated from early on as an alternative to account for the inherent non-linearities of the human body. For example, various studies suggested that geometric representations play a key role in the brain, e.g., to encode perceptual spaces~\cite{Koenderink99:BrainGeometricEngine} and sensorimotor relationships~\cite{Pellionisz85:FunctionalGeometriesCNS,Habas20:GeometriesInCerebellum}.
Geometric methods have also been exploited to better understand the underlying mechanisms of motor control. Handzel and Flash~\cite{Flash99:GeometricMethodsHumanMotorControl} were early to apply tools from differential geometry to account for the intrinsic nonlinear structure of multi-linked systems and the resulting nonlinear couplings between their associated degrees of freedom (DoFs). In particular, the authors analyzed hand trajectories in task space under the lens of affine and equi-affine geometries~\cite{Flash99:GeometricMethodsHumanMotorControl, Flash07:EquiAffineGeometry}, and later proposed a model combining these different geometries which unifies several invariant properties of human motions~\cite{Flash18:MixtureOfGeometryModel}.

The aforementioned works focused on resolving the redundancy present in task space, or in other words, on selecting optimal end-effector trajectories. However, as previously mentioned, the CNS also has to cope with the redundancy of the highly-nonlinear human configuration space. In this context, Riemannian geometry provides suitable tools to describe the phenomena occurring in nonlinear spaces. As shown by Bullo and Lewis~\cite{Bullo05:GeometricControl}, the configuration space of any multi-linked mechanical system can be identified with a Riemannian manifold, i.e., a smooth curved geometric space incorporating the structural and inertial characteristic of the system. On this basis, Biess et al.~\cite{Biess07:ComputationalModelPointing} suggested that arm movements are produced by following \emph{geodesics}, i.e., minimum-muscular-effort or equivalently shortest paths, in the configuration space manifold. 
Importantly, they also showed that the spatial and temporal parts of movements are decoupled, and thus can be determined separately. Their model successfully predicted the joints and hand trajectories of point-to-point motions involving 4 DoFs of the human arm. Planar reaching motions involving 2 DoFs of the arm were also investigated under the lens of Riemannian geometry in~\cite{Sekimoto09:RiemannianDistanceBasedHumanMotion}. In their follow-up work~\cite{Biess11:MovementDynamicsOptimizationInvariance}, Biess et al. reconciled the minimum-jerk and minimum-torque-change models by proving their mathematical equivalence when derived within the Riemannian framework. Their theory was further generalized to the entire human body by Neilson et al.~\cite{Neilson15:GeodesicSynergyHypothesis}, who introduced the so-called \emph{geodesic synergy hypothesis}. Specifically, they described geodesics in the configuration space as movement synergies --- coherent co-activation of motor signals~\cite{dAvella03:MuscleSynergies} --- which are selected by the CNS to overcome the redundancy at the joint level. They additionally conjectured that complex whole-body motions result from a composition of geodesic synergies. However, the hypotheses of~\cite{Neilson15:GeodesicSynergyHypothesis} remain to be experimentally validated.

Inspired by insights from~\cite{Biess07:ComputationalModelPointing,Neilson15:GeodesicSynergyHypothesis}, we hypothesize that human motions are planned as \emph{sequences of geodesic synergies}, i.e., coordinated joint movements achieved with minimum muscular effort. 
Specifically, we model the spatial and temporal aspects of human motion planning using tools from Riemannian geometry similarly to~\cite{Biess07:ComputationalModelPointing,Neilson15:GeodesicSynergyHypothesis} (see Section~\ref{sec:Background} for a short background) and introduce a novel segmentation method which decomposes motions into geodesic synergies (Section~\ref{sec:SequencesGeodesicSynergies}). 
We subsequently validate our hypothesis by providing a complete analysis of various everyday human arm motions, thus investigating the plausibility of our model beyond point-to-point~\cite{Biess07:ComputationalModelPointing} and point-to-manifold~\cite{Berret2011:PointToManifoldAnalysis} movements. 
Our analysis not only shows that geodesic paths account for observed hand paths, arm postures, and speed profiles along the segmented synergies, but also validates that complete motions can indeed be reconstructed as sequences of geodesic synergies (Section~\ref{sec:MotionAnalysis}). Finally, we exploit our Riemannian motion planning framework to transfer, or retarget~\cite{Gleicher98:MotionRetargeting}, human movements to humanoid robots. As explained in Section~\ref{sec:MotionTransfer}, human arm motions are segmented into sequences of geodesic synergies, which are then reproduced as geodesic paths in the robot configuration space. 

The contributions of this paper are threefold: \emph{(i)} we provide a thorough analysis of complex human arm movements viewed as sequences of geodesic synergies; \emph{(ii)} in doing so, we propose a novel Riemannian motion segmentation method; and \emph{(iii)} we introduce a novel perspective on the motion retargeting problem by demonstrating that motions can be transferred as sequences of geodesic synergies. A video accompanying the paper is available at \url{https://sites.google.com/view/riemannian-analysisretargeting/}.

\section{THEORETICAL BACKGROUND}
\label{sec:Background}
In this section, we introduces the mathematical tools that are later exploited to analyze human motions under the lens of Riemannian geometry and to retarget them to robots. 
We refer the interested reader to, e.g.,~\cite{DoCarmo92:RiemannianGeometry,Lee18:RiemannianManifolds}, and to~\cite{Bullo05:GeometricControl} for in-depth introductions to Riemannian geometry, and geometry of mechanical systems, respectively.

\subsection{Riemannian Geometry of Mechanical Systems}
\label{subsec:GeomMechanicalSys}
A $n$-dimensional manifold $\manifold$ is a topological space which is locally Euclidean. In other words, each point in $\manifold$ has a neighborhood which is homeomorphic to an open subset
of the $n$-dimensional Euclidean space $\euclideanspace^n$, also called a chart. The manifold $\manifold$ is smooth if differentiable transitions between charts can be defined. 
The configuration space $\configmanifold$ of a multi-linked mechanical system can be viewed as a smooth manifold with a simple global chart. Points on this manifold correspond to different joint configurations $\jointposition\in\configmanifold$ \footnote{For floating base systems, $\jointposition$ can be augmented with the base pose.}.

Motions of mechanical systems are obtained through smooth changes of joint configurations, i.e., by following smooth trajectories $\jointposition(t)$ in the configuration manifold $\configmanifold$. In general, trajectories on a smooth manifold $\manifold$ are represented geometrically by one-dimensional parametric curves $\jointposition(t)$ with $\jointposition: [a, b] \to \manifold$, and $a,b\in\euclideanspace$. The time differential $\frac{\partial \jointposition}{\partial t}|_{t=a}=\jointvelocity$ is a velocity vector tangent to the manifold at $\jointposition(a)=\jointposition_0$. Namely, it belongs to the tangent space $\tangentspace{\jointposition_0}$ which is the set of the differentials at $\jointposition_0$ of all smooth curves on $\mathcal{M}$ passing through $\jointposition_0$.
The disjoint union of all tangent spaces $\tangentspace{\jointposition}$ forms the tangent bundle $\tangentspace{}$.

A Riemannian manifold is a smooth manifold equipped with a Riemannian metric, i.e., a smoothly-varying inner product acting on $\tangentspace{}$. Given a choice of local coordinates, the Riemannian metric is represented as a symmetric positive-definite matrix $\bm{G}(\jointposition)$, called a metric tensor\footnote{The Euclidean manifold corresponds to $\bm{G}(\jointposition)=\bm{I}$ $\forall \jointposition\in\mathcal{M}$.}, which depends smoothly on $\jointposition\in\manifold$.
The configuration manifold $\configmanifold$ of mechanical systems can be endowed with the so-called \emph{kinetic-energy} metric~\cite{Bullo05:GeometricControl}. Specifically, the metric tensor $\bm{G}(\jointposition)$ is equal to the mass-inertia matrix of the system at the configuration $\jointposition\in\configmanifold$. In that sense, the mass-inertia matrix, i.e., the Riemannian metric, curves the space so that the configuration manifold accounts for the nonlinear inertial properties of the system. Consequently, the Riemannian metric leads to local, nonlinear expressions of inner products and angles.
Specifically, the Riemannian inner product between two velocity vectors $\bm{u}$, $\bm{v}\in\tangentspace{\jointposition_0}$ at $\jointposition_0\in\manifold$ is formulated as a function of the metric tensor
\begin{equation}
    \innerprod{\jointposition_0}{\bm{u}}{\bm{v}} = \langle \bm{u}, \bm{G}(\jointposition_0)\bm{v} \rangle
     \;= \bm{u}^\trsp \bm{G}(\jointposition_0)\bm{v}.
    \label{Eq:InnerProduct}
\end{equation}
Moreover, the Riemannian norm is $\norm{\jointposition_0}{\bm{v}} = \sqrt{\innerprod{\jointposition_0}{\bm{v}}{\bm{v}}}$, and the angle $\theta\in[0, \pi]$ between the two vectors $\bm{u}$, $\bm{v}$ satisfies
\begin{equation}
    \cos{\theta} =
    \frac{\innerprod{\jointposition_0}{\bm{u}}{\bm{v}}}{\norm{\jointposition_0}{\bm{u}}\norm{\jointposition_0}{\bm{v}}}.
    \label{Eq:CosAngle}
\end{equation}
The notions of curve length and energy also need to be adapted to the Riemannian setting, as explained next.

\subsection{Geodesics}
\label{subsec:Geodesics}
As stated by Newton's first law of motion, the kinetic energy of any mechanical system is conserved in the absence of external forces. This implies that the mechanical system follows minimum-energy trajectories unless acted on by an external force. For Euclidean systems, e.g., point-mass particles, these trajectories correspond to move along straight lines at a constant velocity. Due to the space curvature induced by the Riemannian metric, minimum-energy trajectories of multi-linked mechanical systems instead follow \emph{geodesics}, i.e., generalization of straight lines on manifolds.

The kinetic energy\footnote{The kinetic energy is generally used as a geometrical quantity on Riemannian manifolds, even when they are not linked to a physical system. For the sake of clarity, our presentation follows a physical point of view.} of a multi-linked mechanical system is the multi-dimensional generalization of the kinetic energy of a particle. For a joint velocity $\jointvelocity\in\configtangentspace{\jointposition}$ at a configuration $\jointposition\in\configmanifold$, it is given by
\begin{equation}
    k = \frac{1}{2} \norm{\jointposition}{\jointvelocity}^2 = \frac{1}{2} \langle \jointvelocity, \bm{G}(\jointposition)\jointvelocity \rangle = \frac{1}{2} \jointvelocity^\trsp \bm{G}(\jointposition)\jointvelocity. 
    \label{Eq:KineticEnergy}
\end{equation}
In the absence of external forces, the total energy along a curve $\jointposition(t)$ is thus obtained as $E = \int_a^b k(t) dt$. The energy function $E$ is minimized by a geodesic curve, which follows from the application of the Euler-Lagrange equations to~\eqref{Eq:KineticEnergy}. Namely, geodesics solve the following system of second-order ordinary differential equations (ODE)
\begin{equation}
    \sum_{j} g_{ij}(\jointposition) \ddot{q}_j + \sum_{jk} \Gamma_{ijk} \dot{q}_j \dot{q}_k = 0,
    \label{Eq:Geodesic}
\end{equation}
where $\sum_{jk} \Gamma_{ijk} \dot{q}_j \dot{q}_k = c_i(\jointposition, \jointvelocity)$ represents the influence of Coriolis forces, $\Gamma_{ijk} = \frac{1}{2} \left( \frac{\partial g_{ij}}{\partial q_k} + \frac{\partial g_{ik}}{\partial q_j} - \frac{\partial g_{jk}}{\partial q_i} \right)$ are the Christoffel symbols of the first kind, and $q_k$ and $g_{ij}$ denote the $k$-th and $i,j$-th component of $\jointposition$ and $\bm{G}$. 
In other words, geodesic trajectories are obtained by applying the joint acceleration $\jointacceleration(t)$ solution of~\eqref{Eq:Geodesic} at each configuration $\jointposition(t)$ with velocity $\jointvelocity(t)$ along the trajectory. Similarly to straight lines in Euclidean spaces, the kinetic energy $k(t)$, and thus the velocity, is conserved along a geodesic. 
Moreover, one can easily prove that curves minimizing the energy function $E$ are also locally minimum-length curves, where the length of a curve is defined on Riemannian manifolds as 
\begin{equation}
    \ell \left(\jointposition(t)\right) = \int_{a}^{b} \norm{\jointposition(t)}{\jointvelocity(t)} dt.
    \label{Eq:LengthCurve}
\end{equation}
Therefore, similarly to straight lines in Euclidean space, geodesics are minimum-energy and minimum-length, constant-velocity curves on Riemannian manifolds.

\subsection{Operations on Riemannian manifolds}
\label{subsec:Operators}
\begin{figure}
\centering
	\includegraphics[width=.38\textwidth]{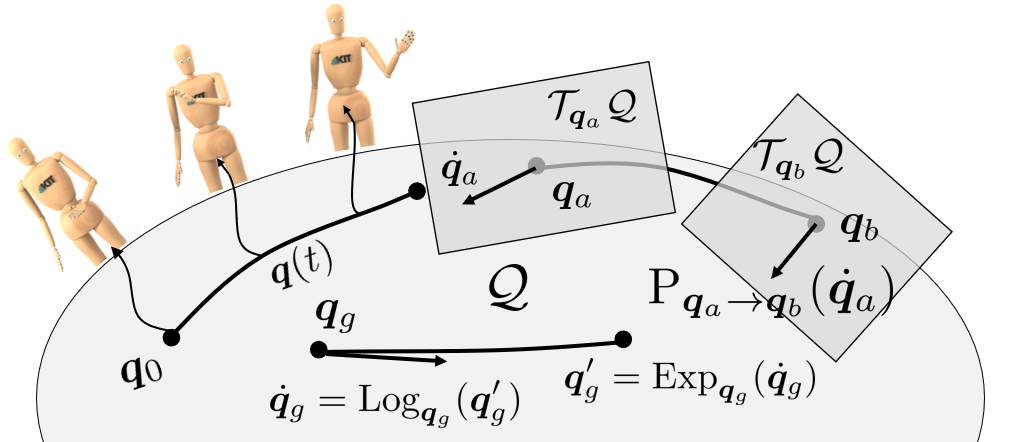}
	\caption{Illustration of the configuration manifold $\configmanifold$, along with the exponential, logarithmic map, and parallel transport. Trajectories on the manifold correspond to smooth changes in joint configurations. Observe that velocities at different points $\jointposition$ exist in disjoint tangent spaces. Thus, parallel transport is required to compare them. }
	\label{Fig:ManifoldOperations}
	\vspace{-0.4cm}
\end{figure}

As previously explained, geodesics are the solution to the linear system of second-order ODEs~\eqref{Eq:Geodesic}, and are completely determined by their initial conditions ${\jointposition}(0)$ and ${\jointvelocity}(0)$. This allows us to define the so-called \emph{exponential map} $\expmap{\jointposition_0}{\jointvelocity_0}$  at any point $\jointposition_0\in\configmanifold$ as the solution at time $t\!=\!1$ of the geodesic ODE~\eqref{Eq:Geodesic} starting at $\jointposition_0$ with velocity $\jointvelocity_0$. This corresponds to solving an initial value problem (IVP). 
Given another point $\jointposition_1\in\configmanifold$, the inverse operation, called the \emph{logarithmic map} $\logmap{\jointposition_0}{\jointposition_1}$, computes the initial velocity $\jointvelocity_0$ such that $\expmap{\jointposition_0}{\jointvelocity_0} = \jointposition_1$. Namely, it solves the boundary value problem (BVP) of~\eqref{Eq:Geodesic} with initial and final conditions $\jointposition(0)=\jointposition_0$ and $\jointposition(1)=\jointposition_1$. Consequently, the logarithmic map is usually harder to compute than the exponential map.

Since joint velocities at different points on the manifold exist in disjoint tangent spaces, they cannot be directly compared and must first be transported into a common tangent space.
Specifically, given a curve $\gamma(t)$ along which to transport a vector $\jointvelocity$, the \emph{parallel transport} operation corresponds to solving the following system of ODEs
\begin{equation}
    \sum_{j} g_{ij}(\jointposition) \ddot{q}_j + \sum_{jk} \Gamma_{ijk} \dot{\gamma}_j \dot{q}_k = 0.
    \label{Eq:ParallelTransport}
\end{equation}
The parallel transport of $\jointvelocity$ from $\configtangentspace{\gamma(a)}$ to $\configtangentspace{\gamma(b)}$ is denoted $\prltrsp{\gamma(a)}{\gamma(b)}{\jointvelocity}$.
Note that~\eqref{Eq:ParallelTransport} is remarkably similar to~\eqref{Eq:Geodesic}: Indeed, the initial velocity $\jointvelocity\in\configtangentspace{\jointposition}$ is parallel-transported along the geodesic it generates. Figure~\ref{Fig:ManifoldOperations} illustrates the aforementioned operations.
In this paper, we first exploit the parallel transport to compare velocities along an observed motion to segment it into geodesic synergies. We then exploit the logarithmic map for planning individual synergies.

\section{HUMAN MOTIONS AS SEQUENCES OF GEODESIC SYNERGIES}
\label{sec:SequencesGeodesicSynergies}
One of the most important characteristics of humans is their ability to move and act in their environment. In that sense, uncovering the underlying mechanisms of human motion generation is paramount for understanding the global functioning of the human brain and body. In this section, we first develop the hypothesis that human motions are planned as sequences of geodesic synergies. We then introduce the corresponding computational model, which is exploited for analyzing real human motions in Section~\ref{sec:MotionAnalysis}.

\subsection{Sequences of geodesic synergies}
\label{subsec:GeodesicSynergies}
Several experiments demonstrated that simple point-to-point, e.g., reaching, motions are successfully predicted by following geodesics in the human configuration manifold endowed with the kinetic-energy metric~\cite{Biess07:ComputationalModelPointing, Sekimoto09:RiemannianDistanceBasedHumanMotion, Biess11:MovementDynamicsOptimizationInvariance}. The corresponding joint coordinations are called \emph{geodesic synergies} or geodesic movement synergies~\cite{Neilson15:GeodesicSynergyHypothesis}. Although single geodesic synergies are sufficient to plan simple short point-to-point movements, they may not allow the planning of complex movements. Indeed, complex movements can be seen as sequences of simpler motion units. For example, when a human waves a hand in greeting, the hand is first lifted, then several left-to-right and right-to-left waving motions are executed before bringing back the arm along the body. As these units account for different parts of the overall movement, they may rather be planned as different geodesic synergies which are then sequenced into a complete motion. 
To validate our hypothesis, we verify if observed human arm motions can be reconstructed as piecewise geodesic.
To do so, two main challenges must be tackled, namely, \emph{(i)} how to segment motions into different geodesic synergies, and \emph{(ii)} how to model the individual synergies. 

\subsection{Riemannian motion segmentation}
\label{subsec:MotionSegmentation}
As explained in Section~\ref{subsec:Geodesics}, the direction of the velocity is conserved along geodesic curves. We here exploit this feature to design a novel Riemannian segmentation method for human motions. Namely, consider that the $g$-th geodesic synergy started at time $t_i^{(g)}$ at the initial configuration ${\jointposition_i^{(g)}\in\configmanifold}$ and initial velocity $\jointvelocity_i^{(g)}\in\configtangentspace{\jointposition_i^{(g)}}$. We assume that the next points $\jointposition_t$ of the observed joint motion belong to the same synergy as long as the velocity direction is conserved. In other words, if the velocities $\jointvelocity_i^{(g)}$ and $\jointvelocity_t$ point to the same direction, $\jointposition_t$ belong to the $g$-th synergy. Otherwise, a new synergy is initialized. Importantly, as $\jointvelocity_i^{(g)}$ and $\jointvelocity_t$ belong to two different tangent spaces, $\jointvelocity_i^{(g)}$ must be parallel transported to $\configtangentspace{\jointposition_t}$ in order to be compared to $\jointvelocity_t$. The proposed Riemannian motion segmentation is summarized in Algorithm~\ref{Algo:MotionSegmentation}.
Note that the segmentation only considers the angle between velocity vectors, as their norm may vary along a geodesic synergy. As detailed next, although each segment follows a geodesic path in the configuration manifold, the speed at which these paths are traversed may not be constant. 

\begin{algorithm}
	\caption{Riemannian motion segmentation}
	\label{Algo:MotionSegmentation}
	\KwIn{Observed human arm trajectory $\{\jointposition_t\}_{t=0}^T$, threshold $\Delta_{\theta}$}
	\KwOut{Segmented synergies with $\{t_i^{(g)}, t_f^{(g)}\}_{g=1}^G$}
	Initialize the first synergy at $t_i^{(1)}=0$ \;
	\While{$t \leq T$}{
	    Parallel transport $\jointvelocity_i^{(g)}$ to $\jointposition_t$ by solving~\eqref{Eq:ParallelTransport}, 
	    \vspace{-0.2cm}
	    \begin{equation*}
	    \text{set }
	        \tilde{\jointvelocity}_i^{(g)} = \prltrsp{\jointposition_i^{(g)}}{\jointposition_t}{\jointvelocity_i^{(g)}}.
	    \end{equation*} \\
	    \vspace{-0.2cm}
	    Compute the angle $\theta$ between $\tilde{\jointvelocity}_i^{(g)}$ and $\jointvelocity_{t}$~\eqref{Eq:CosAngle}.\\
		\If{$\theta > \Delta_{\theta}$}{
		Start a new synergy: $t_f^{(g)}\!=\!t$, $t_i^{(g+1)}\!=\!t\!+\!dt$.
		}
	}
\end{algorithm}

\vspace{-0.2cm}
\subsection{Computation of geodesic synergies}
\label{subsec:GeodesicSynergyComputation}
Given the Riemannian segmentation of an observed human motion, we are then interested in planning the individual geodesic synergies that constitute the different segments. To do so, we follow the computational models of~\cite{Biess07:ComputationalModelPointing, Neilson15:GeodesicSynergyHypothesis} to determine the spatial and temporal aspects of each synergy.

More specifically, the \emph{spatial} aspect of a geodesic synergy refers to the geometric path followed in the configuration manifold $\configmanifold$, which can also be seen as the specific coordination between the different joints along the movement. According to the hypothesis that human arm motions follow minimum-muscular-effort paths, each synergy spatially corresponds to a geodesic in $\configmanifold$. Therefore, given the observed initial and final joint configurations $\jointposition_i^{(g)}, \jointposition_f^{(g)}\in\configmanifold$, the spatial path $\jointposition(s)$ of the $g$-th geodesic synergy is obtained by computing the logarithmic map from $\jointposition_i^{(g)}$ to $\jointposition_f^{(g)}$. 
As previously mentioned, the velocity of the motion is constant along a geodesic. However, most human movements require some acceleration, e.g., when moving away from a still posture.
As shown in~\cite{Biess07:ComputationalModelPointing}, such accelerations can be introduced by modifying the speed at which the spatial path $\jointposition(s)$ is traversed, i.e., by reparametrizing the geodesic as $\jointposition(s(t))$. This is referred to as the \emph{temporal} aspect of the geodesic synergy. It is important to notice that this reparametrization does not change the direction of the velocity along the geodesic --- the spatial path remains the same ---, but instead modifies its norm $\norm{\jointposition(s(t))}{\jointvelocity(s(t))}$, which may not be constant anymore\footnote{The velocity norm remains constant only if $s$ is an affine function of $t$.}. 
The temporal path of the $g$-th geodesic synergy is obtained by setting its initial and final velocities to the observed velocities $\jointvelocity_i^{(g)},\jointvelocity_f^{(g)}$, thus accounting for possible accelerations along the synergy.
Following~\cite{Neilson15:GeodesicSynergyHypothesis}, we define the temporal path $s(t)$ as the minimum-energy time course along the geodesic. This corresponds to minimizing the integrated squared acceleration
\begin{equation}
    \int_{t=0}^{T^{(g)}} \ddot{s}^2(t) dt,
    \label{Eq:GeodesicTemporalPath}
\end{equation}
subject to the boundary conditions
$s(0)=0$, $s(T^{(g)})=\ell^{(g)}$, $\dot{s}(0) = \norm{\jointposition_i^{(g)}}{\jointvelocity_i^{(g)}}$, $\dot{s}(T^{(g)}) = \norm{\jointposition_f^{(g)}}{\jointvelocity_f^{(g)}}$,
where $\ell^{(g)}$ and $T^{(g)}=t_f^{(g)} - t_i^{(g)}$ denote the length~\eqref{Eq:LengthCurve} and total duration of the $g$-th geodesic synergy, respectively.

\section{RIEMANNIAN MOTION ANALYSIS}
\label{sec:MotionAnalysis}
In this section, we investigate if human arm motions may be planned as sequences of geodesic synergies. To do so, we present a detailed analysis of several everyday human arm motions based on the computational model of Section~\ref{sec:SequencesGeodesicSynergies}.

\subsection{Data description and processing}
\label{subsec:DataDescription}
For our analysis, we consider several human movements from the KIT whole-body human motion database~\cite{Mandery2016b:MotionDatabase}\footnote{\url{https://motion-database.humanoids.kit.edu/}}, namely  one $\mathsf{reaching}$, one $\mathsf{throwing}$, one $\mathsf{pointing}$ (successively up, horizontally, and down), and two $\mathsf{waving}$ (in greeting) motions\footnote{Motions identified in the database as take\_book\_from\_shelf\_right\_arm\_01, throw\_right01, point\_at\_right03, wave\_left01, and waving\_neutral04.}.
As all these movements mostly involve the joints of one arm, and thus we restrict our analysis to the $7$-DoFs of the left or right human arm.

All motions in the database were recorded with a marker-based optical motion capture system. The motions were then mapped to the Master Motor Map (MMM) model~\cite{Mandery2016b:MotionDatabase} --- a reference model of the human body including statistical, kinematic, and dynamic properties, such as the center of mass and inertia tensor per segment adapted from~\cite{Leva1996:MotionDatabase}.
As in~\cite{Mandery2016b:MotionDatabase}, the mapping from the markers to the joint configurations of the MMM model was achieved by minimizing the error between recorded and virtual marker positions obtained on the model with forward kinematics. In comparison to the previous work, criteria inspired by~\cite{Rakita18:MotionDatabase} were included in order to reduce the joint velocity, acceleration, and jerk. The optimization was conducted with the sequential quadratic programming algorithm of NLopt~\cite{Johnson2011:MotionDatabase}. The obtained joint trajectories are considered as ground truth for our analysis. The joint velocities were computed as the analytical derivative of a second-order approximation of the joint angles over time obtained with a Savitzky–Golay filter (window-length of 21 samples).
The mass-inertia matrix $\bm{G}(\jointposition)$ of the human was computed by applying the composite rigid body algorithm of RBDL~\cite{Felis2016:MotionDatabase} on the MMM model.
The algorithm was adapted to calculate the first and second derivatives of the mass-inertia matrix, required to solve the system of ODEs~\eqref{Eq:Geodesic}. 

\begin{figure*}[tbp]
	\centering
	\begin{subfigure}[b]{0.185\textwidth}
		\adjustbox{trim={.15\width} {.37\height} {0.4\width} {.2\height},clip}{\includegraphics[width=1.72\textwidth]{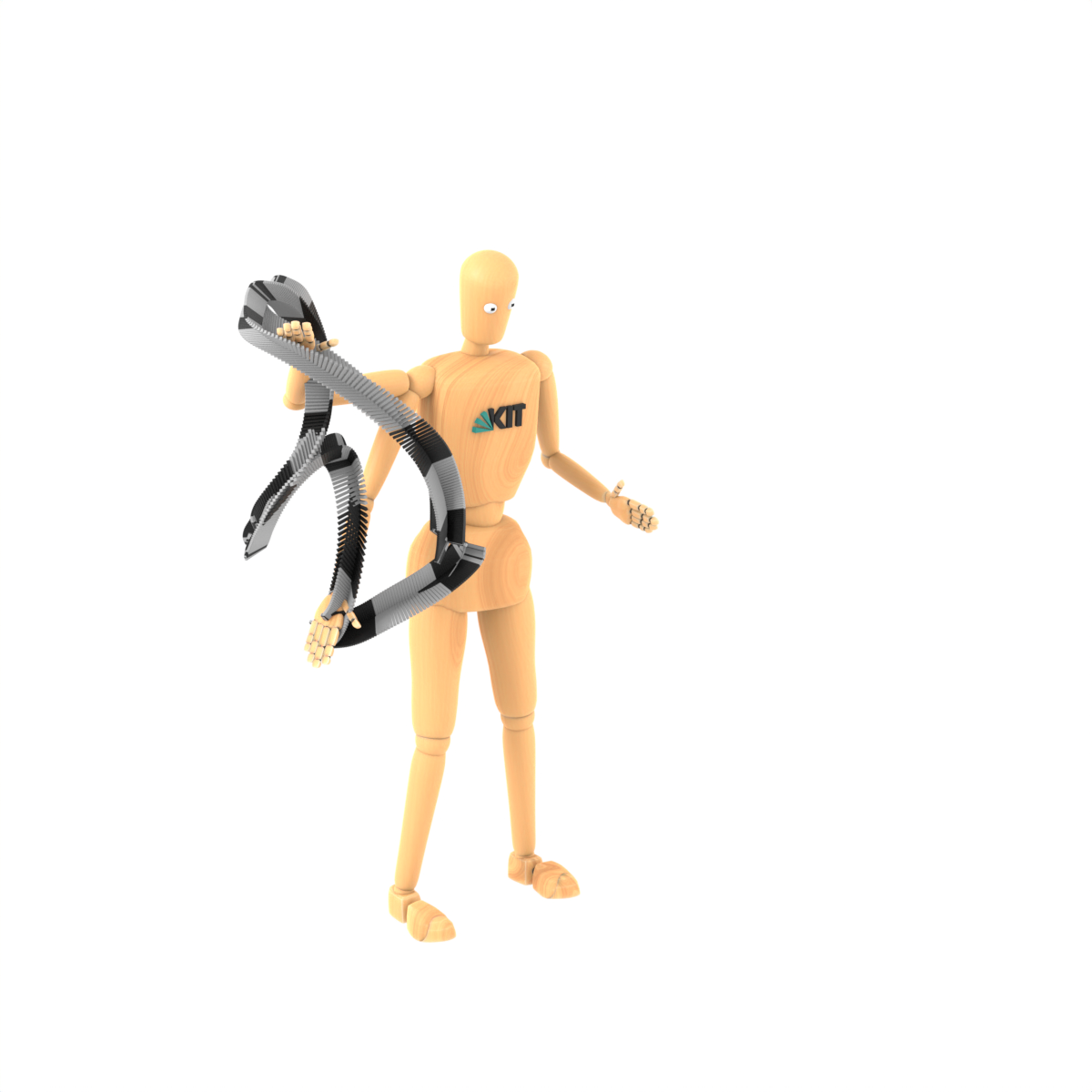}}
		\adjustbox{trim={.15\width} {.37\height} {0.4\width} {.2\height},clip}{\includegraphics[width=1.72\textwidth]{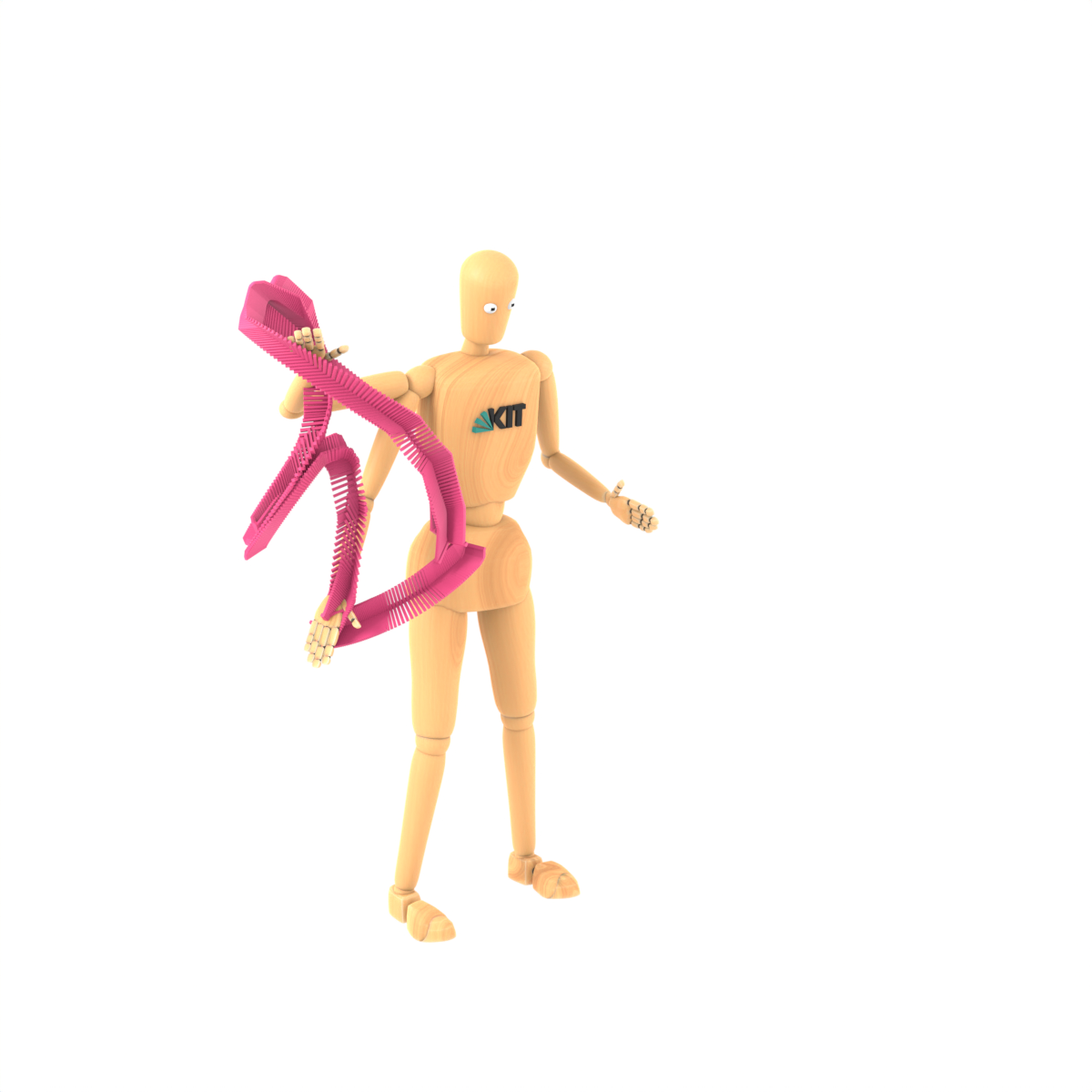}}
		\adjustbox{trim={.15\width} {.37\height} {0.4\width} {.2\height},clip}{\includegraphics[width=1.72\textwidth]{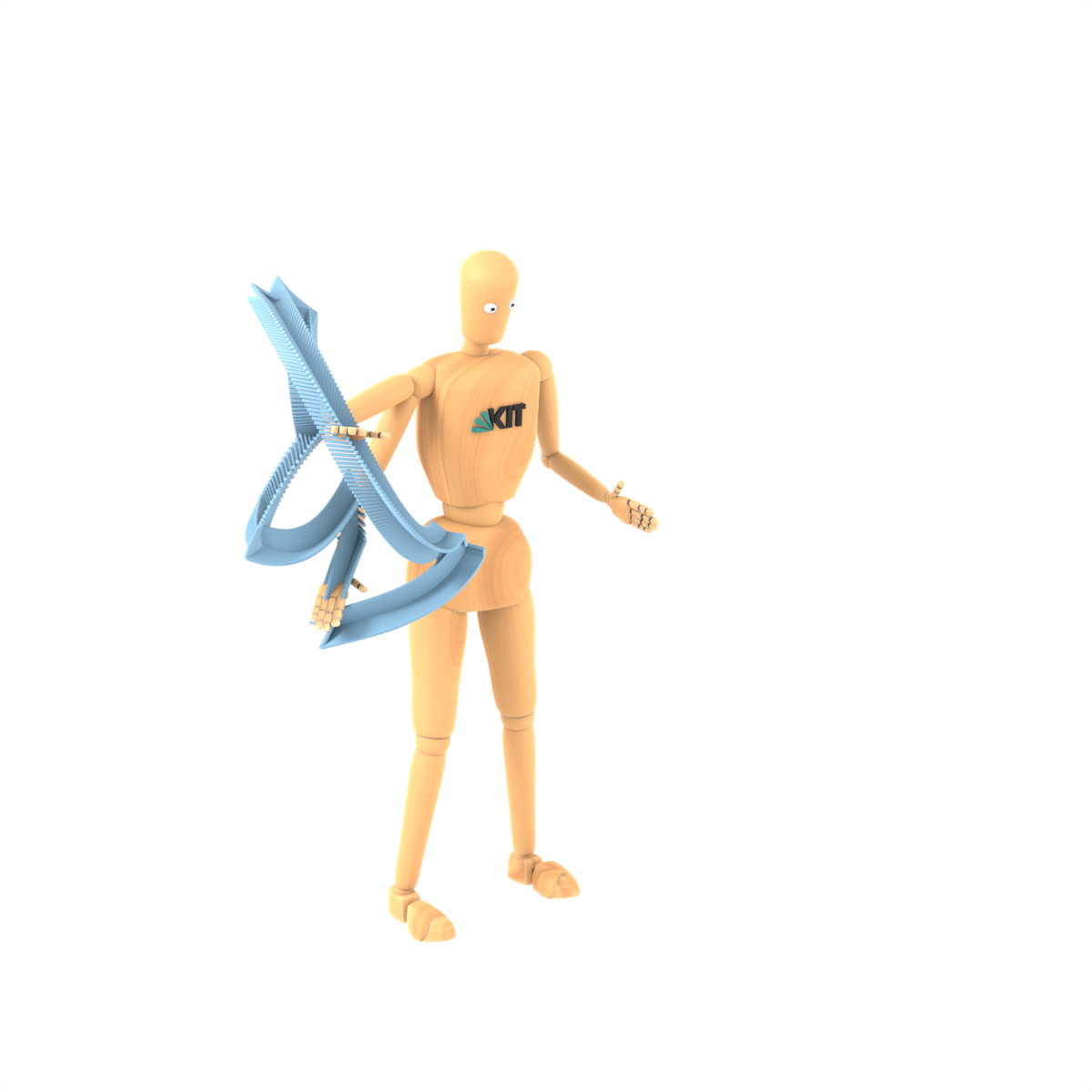}}
		\adjustbox{trim={.15\width} {.37\height} {0.4\width} {.2\height},clip}{\includegraphics[width=1.72\textwidth]{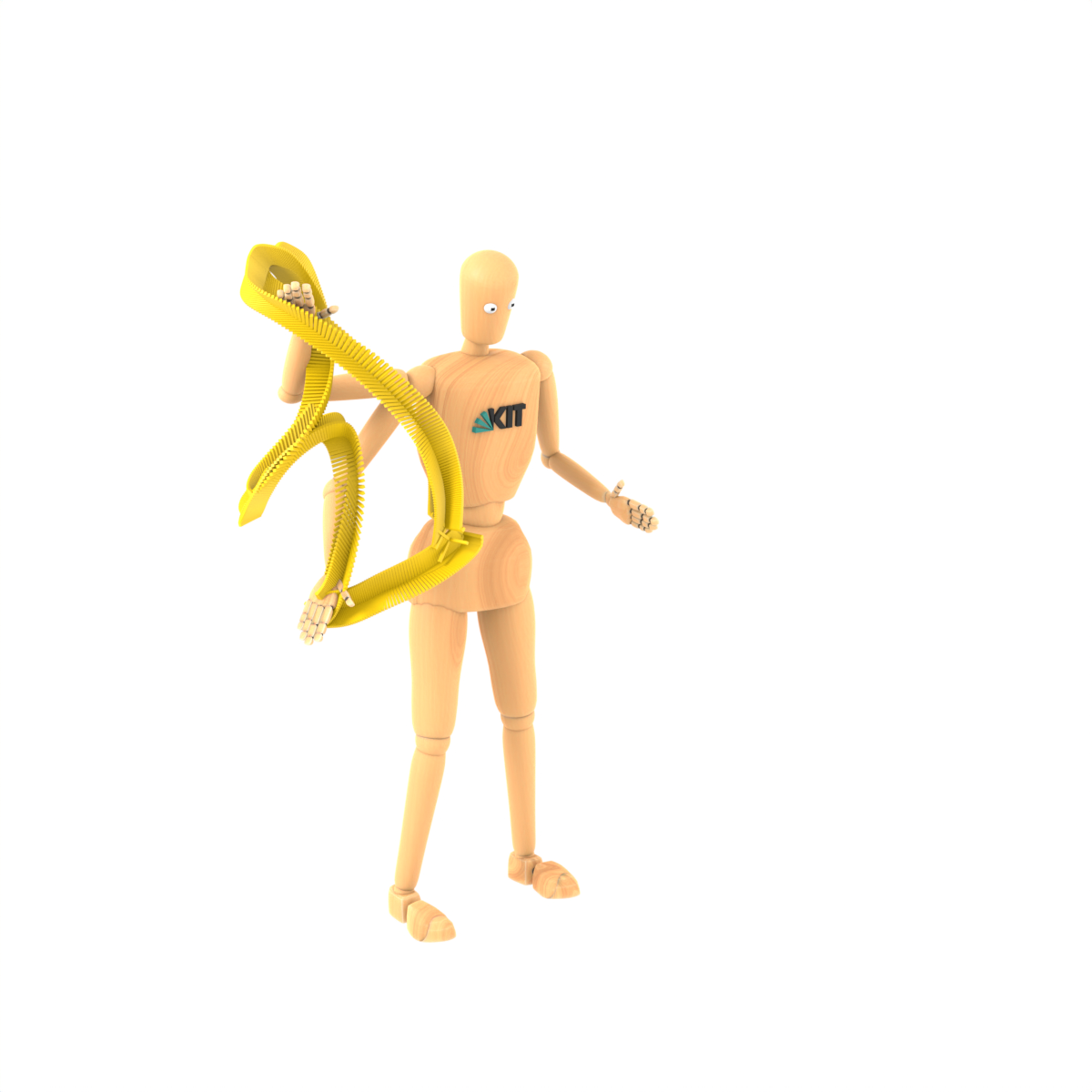}}
		\caption{$\mathsf{Pointing}$ motion}
		\label{subFig:PointingMMM}
	\end{subfigure}
	\begin{subfigure}[b]{0.3\textwidth}
		\includegraphics[width=0.95\textwidth,trim={0cm 0cm 0cm 0cm},clip]{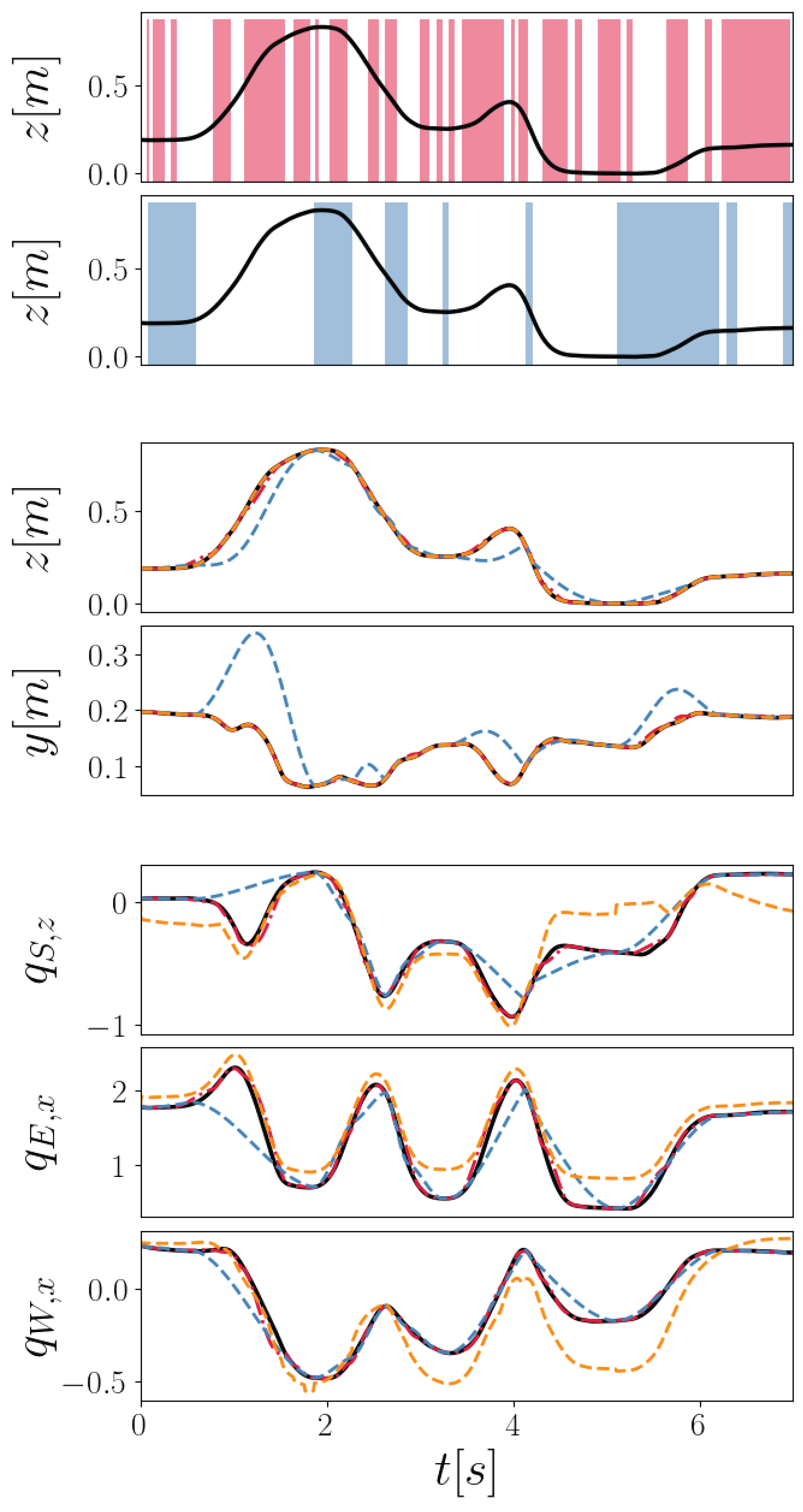}
		\caption{$\mathsf{Pointing}$ segmentations and reproductions}
		\label{subFig:PointingGraphs}
	\end{subfigure}
	\begin{subfigure}[b]{0.185\textwidth}
		\adjustbox{trim={.1\width} {.43\height} {0.3\width} {.12\height},clip}{\includegraphics[width=1.65\textwidth]{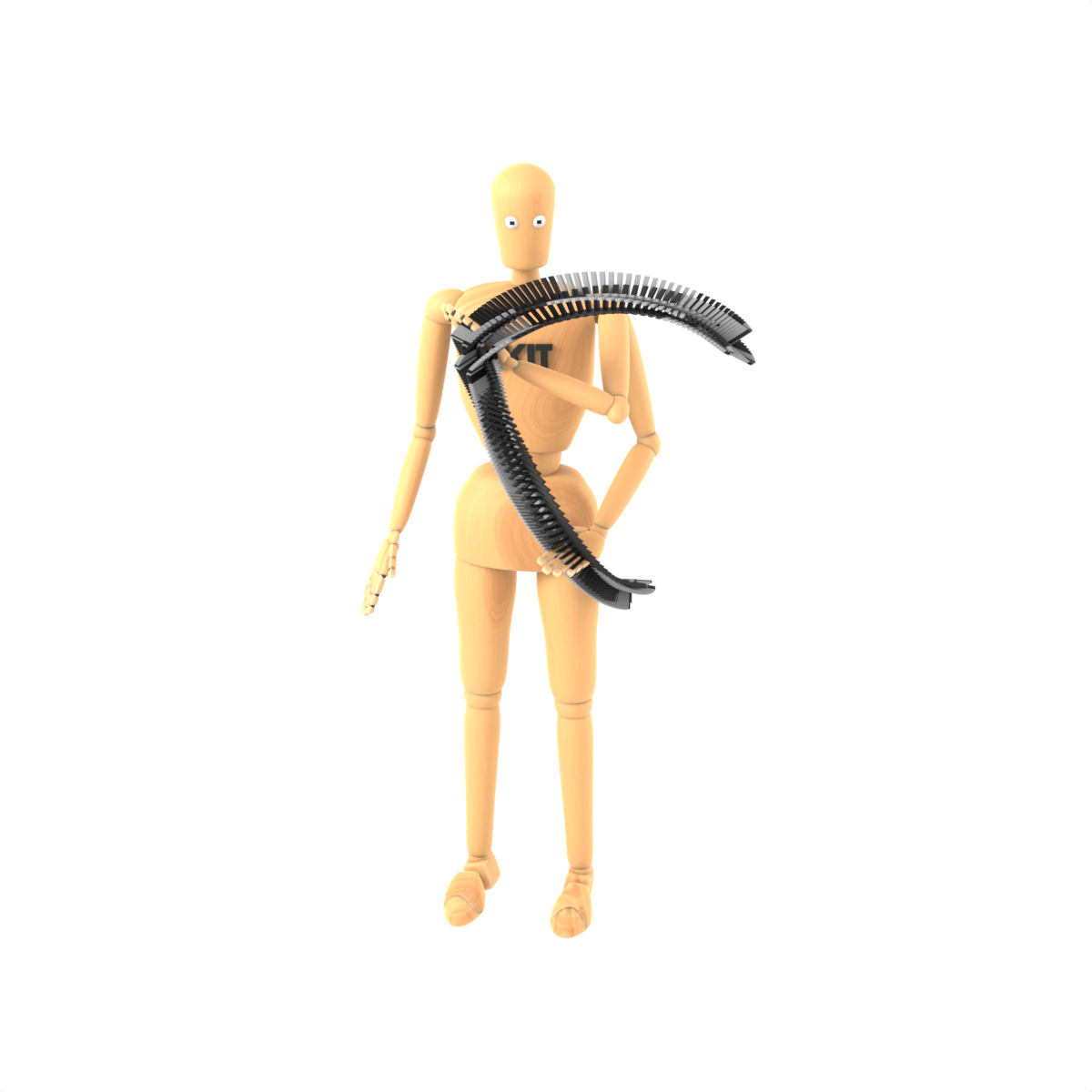}}
		\adjustbox{trim={.1\width} {.43\height} {0.3\width} {.12\height},clip}{\includegraphics[width=1.65\textwidth]{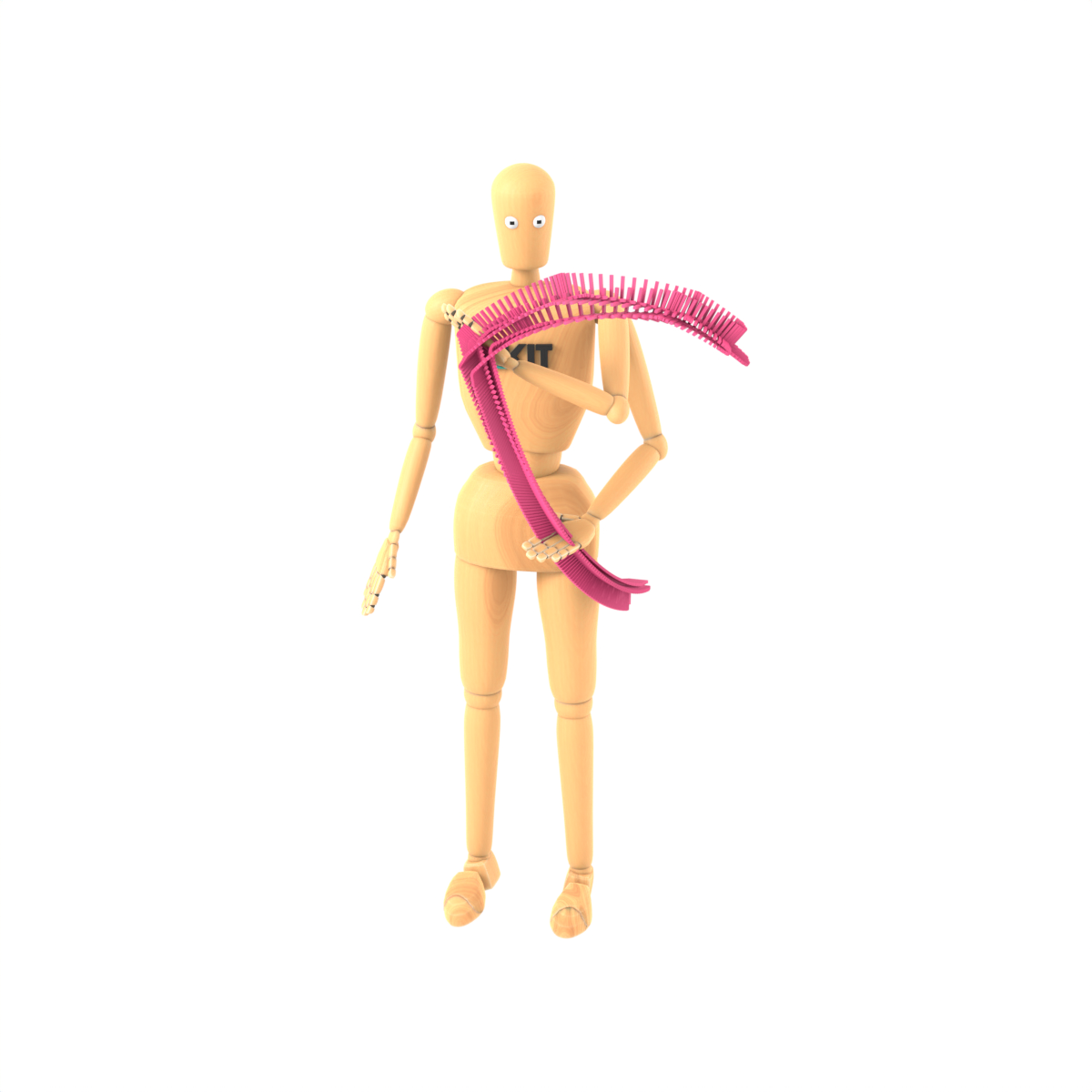}}
		\adjustbox{trim={.1\width} {.43\height} {0.3\width} {.12\height},clip}{\includegraphics[width=1.65\textwidth]{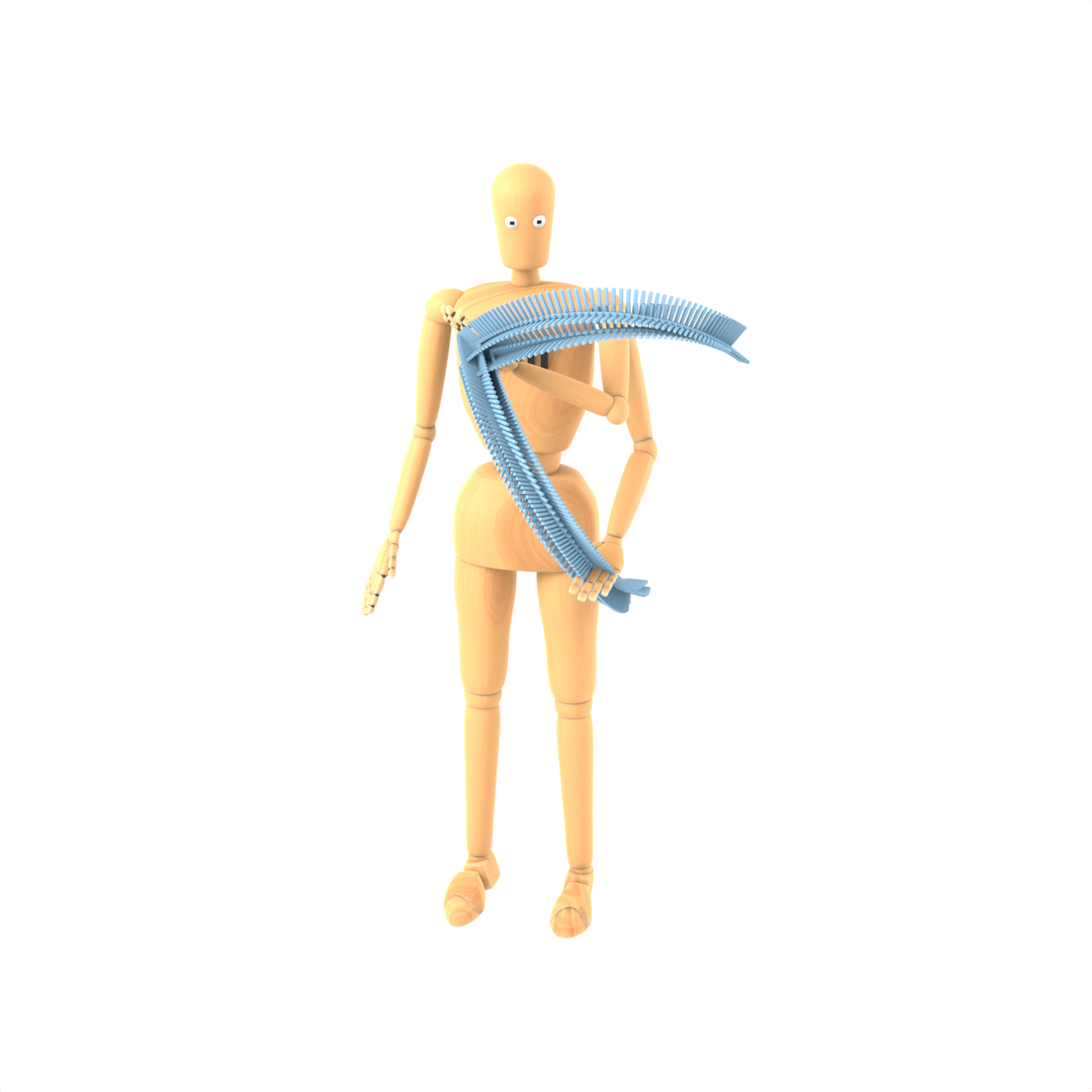}}
		\adjustbox{trim={.1\width} {.43\height} {0.3\width} {.12\height},clip}{\includegraphics[width=1.65\textwidth]{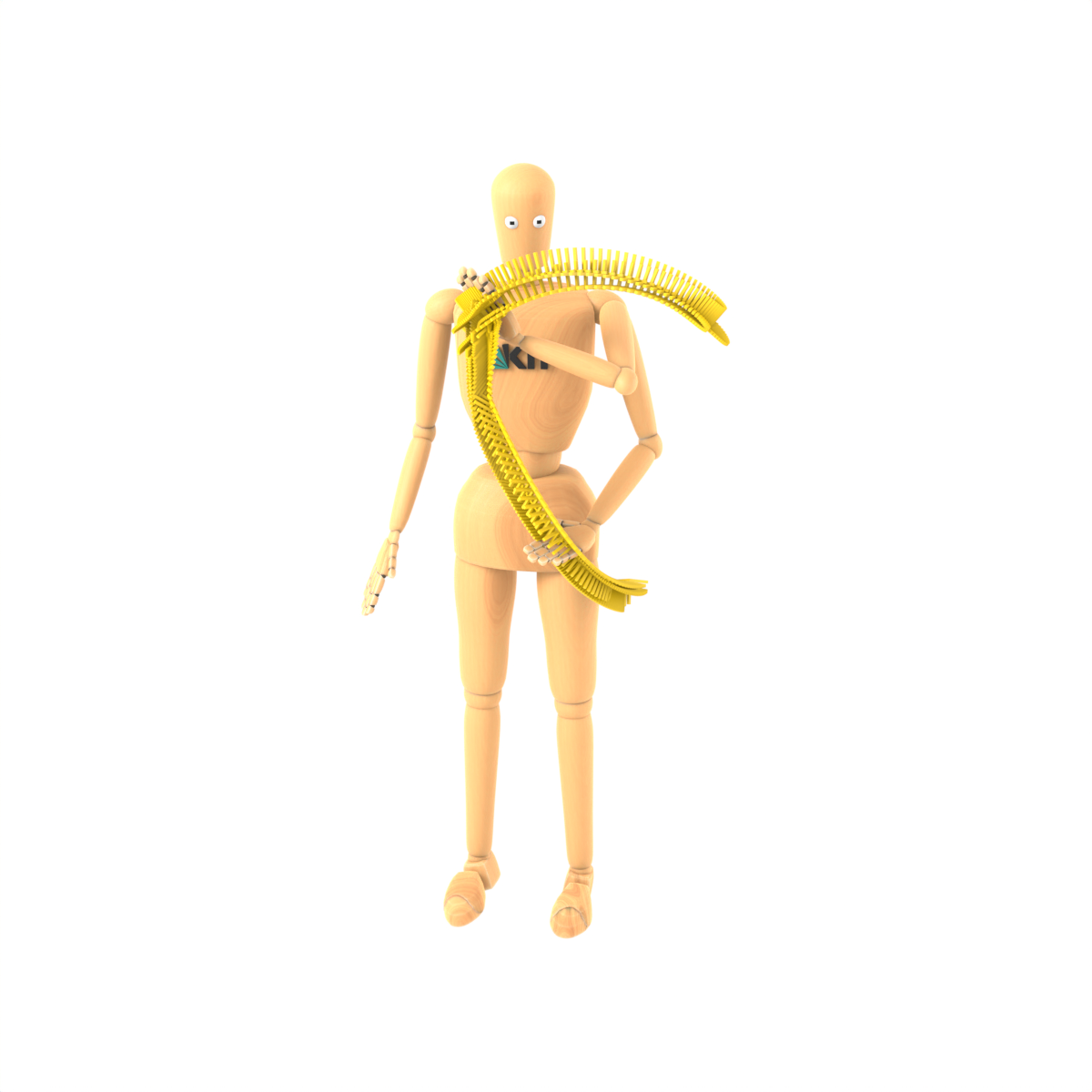}}
		\caption{$\mathsf{Waving}$ motion}
		\label{subFig:WavingMMM}
	\end{subfigure}
	\begin{subfigure}[b]{0.3\textwidth}
		\includegraphics[width=0.95\textwidth,trim={0cm 0cm 0cm 0cm},clip]{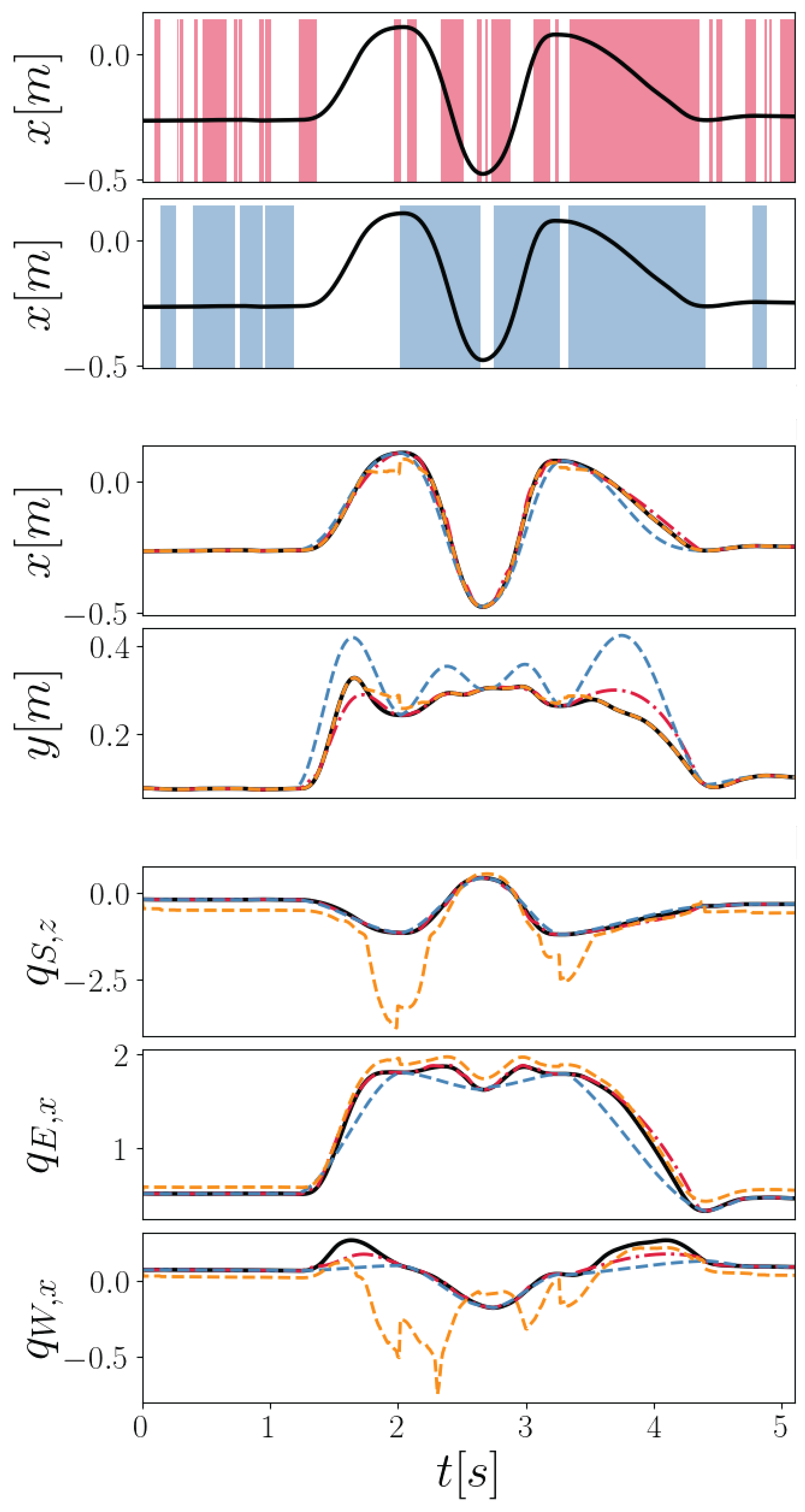}
		\caption{$\mathsf{Waving}$ segmentations and reproductions}
		\label{subFig:WavingGraphs}
	\end{subfigure}
	\caption{Segmentation and trajectories for \emph{(a)}-\emph{(b)} $\mathsf{pointing}$ and \emph{(c)}-\emph{(d)} short $\mathsf{waving}$ motions. \emph{(a,c) - top to bottom}: Ground-truth human motion, and reproductions with the Riemannian, Euclidean, and IK models. The posture of the model is shown at times \emph{(a)} $t=\{1.4, 4.6\}$s and \emph{(c)} $t=\{2,4\}$s. The Riemannian synergies are depicted by alternated colors on the ground-truth motion. \emph{(b,d) - 2 top rows}: Riemannian (\redrectangle \redrectangle) and Euclidean (\bluerectangle \bluerectangle) segmentations with the observed hand trajectory (\blackline) along a relevant axis. Segments are denoted by alternated colors. \emph{(b,d) - 2 middle rows}: Ground-truth (\blackline) and predicted hand paths with the Riemannian (\redline), Euclidean (\blueline), and IK (\orangeline) models along two axes. \emph{(b,d) - 3 bottom rows}: Ground-truth and predicted trajectories for the shoulder, elbow and wrist joints $q_{S,z}$, $q_{E,x}$, $q_{W,x}$.}
	\label{Fig:AnalysisResults}
	\vspace{-0.4cm}
\end{figure*}

\subsection{Analysis}
\label{subsec:Analysis}
Here we examine the predictions of the Riemannian model and compare them with the results of a Euclidean approach. In the latter, the motions are segmented using a classical zero-velocity crossing approach~\cite{Ajo02:VelocityCrossingSeg} in which a new segment is added if more than 3 joint velocities cross 0 within a window of $50$ms.
The individual segments are then predicted by assuming a Euclidean metric on the configuration space, for which minimum-energy trajectories are straight lines temporally traversed as in~\eqref{Eq:GeodesicTemporalPath}. 

The Riemannian segmentation is illustrated in Figs.~\ref{subFig:PointingMMM} and~\ref{subFig:WavingMMM} (top) on the ground-truth $\mathsf{pointing}$ and short $\mathsf{waving}$ motions, respectively. We observe that the core parts of the motions, e.g., lifting, putting down the hand, or waving left-to-right, are usually planned as one or two long geodesic synergies, while the transitions are encoded by several small ones. Interestingly, this relates to the results in~\cite{Sekimoto09:RiemannianDistanceBasedHumanMotion}, which suggest that point-to-point motions are further apart from geodesics at the beginning and end of the movement than during its core part.
This can be explained via the Riemannian segmentation, which shows that a larger quantity of smaller geodesic synergies are required at the transitions, including at the start and end of the motion.
As shown by Figs.~\ref{subFig:PointingGraphs} and~\ref{subFig:WavingGraphs} (2 top rows), the Riemannian approach results into more segments than the Euclidean one. Interestingly, there is usually a correspondence between long geodesic synergies and long Euclidean segments. This is particularly obvious, e.g., for the $\mathsf{waving}$ motion (Fig.~\ref{subFig:WavingGraphs}) from $t\!=\!1$ to $4$s.

The configurations predicted by the Riemannian and Euclidean models for a shoulder, an elbow, and a wrist joint are depicted in Figs.~\ref{subFig:PointingGraphs} and~\ref{subFig:WavingGraphs} (3 bottom rows) for the $\mathsf{pointing}$ and short $\mathsf{waving}$ motions, respectively. As a baseline, we compute the joint configurations from the ground-truth task-space trajectories using a standard Jacobian-based velocity controller, which we refer to as IK model.
The corresponding hand trajectories are displayed for the most relevant dimensions in the 2 middle rows, and illustrated along with the MMM model in Figs.~\ref{subFig:PointingMMM} and~\ref{subFig:WavingMMM}.

We observe that both motions are successfully reconstructed with sequences of geodesic synergies. Importantly, the Riemannian model accurately predicts the spatial (Fig.~\ref{subFig:PointingMMM},~\ref{subFig:WavingMMM}) and temporal (Fig.~\ref{subFig:PointingGraphs},~\ref{subFig:WavingGraphs}) properties of the motions at both joint- and task-space levels. In contrast, the Euclidean model predicts straight lines in joint space, which often do not match the observed curved joint trajectories. Consequently, the resulting hand paths often differ from the ground-truth ones. This is particularly visible, e.g., in the top and left of Fig.~\ref{subFig:PointingMMM} where the human points up and horizontally.
Notice that similar observations have been made for 4-DoFs short point-to-point motions, which were better predicted with single geodesics than with Euclidean paths~\cite{Biess07:ComputationalModelPointing}. Importantly, the results presented in this section generalize these observations to complex motions composed of several geodesic synergies. It is worth emphasizing that the Riemannian model also outperforms the predictions of the Euclidean model when both segmentations are similar, see e.g., Fig.~\ref{subFig:WavingGraphs} for $t\!=1.3$ to $2$s and $t\!=3.4$ to $4.4$s.
As expected, the hand trajectories obtained with IK coincide precisely with the observed ones. However, the underlying joint postures often differ from the ground-truth motion. Indeed, due to the redundancy of the human arm, the same hand poses can be obtained from different joint trajectories.

The aforementioned observations are validated quantitatively for all the considered motions in Tables~\ref{Tab:ErrorsJointAngles} and~\ref{Tab:ErrorsPoses}.
We observe that the Riemannian model reliably predicts the observed human arm postures and hand trajectories, and thus always outperforms the Euclidean model.

\begin{table}[t]
	\renewcommand*{\arraystretch}{1.2}
	\caption{Mean joint-angle error w.r.t. ground-truth motions. The means are computed over the complete trajectories.}
	\label{Tab:ErrorsJointAngles}
	\begin{center}
		\begin{tabular}{c|c|c|c|}
			& Riemannian & Euclidean & IK\\
			\hline
			$\mathsf{reaching}$ & $\mathbf{0.036}$ & $0.427$ & $0.393$\\
			$\mathsf{throwing}$ & $\mathbf{0.029}$ & $0.120$ & $0.864$\\
			$\mathsf{pointing}$ & $\mathbf{0.047}$ & $0.188$ & $0.488$\\
			$\mathsf{waving}$ (short) & $\mathbf{0.049}$ & $0.095$ & $0.470$\\
			$\mathsf{waving}$ (long) & $\mathbf{0.038}$ & $0.074$ & $0.393$\\
			\hline
		\end{tabular}
	\end{center}
\end{table}

\begin{table}[t]
	\renewcommand*{\arraystretch}{1.2}
	\caption{Mean hand-pose error w.r.t. ground-truth motions. IK is not considered as it perfectly tracked the ground-truth hand poses.}
	\label{Tab:ErrorsPoses}
	\begin{center}
		\begin{tabular}{c|c|c|c|c|}
		    & \multicolumn{2}{c|}{Position} & \multicolumn{2}{c|}{Orientation} \\
			& Riem. & Eucl. & Riem. & Eucl.\\
			\hline
			$\mathsf{reaching}$ & $\mathbf{0.009}$ & $0.094$ & $\mathbf{0.014}$ & $0.181$ \\
			$\mathsf{throwing}$ & $\mathbf{0.007}$ & $0.035$ & $\mathbf{0.012}$ & $0.053$ \\
			$\mathsf{pointing}$ & $\mathbf{0.008}$ & $0.047$ & $\mathbf{0.015}$ & $0.076$ \\
			$\mathsf{waving}$ (short) & $\mathbf{0.012}$ & $0.029$ & $\mathbf{0.017}$ & $0.040$ \\
			$\mathsf{waving}$ (long) & $\mathbf{0.009}$ & $0.026$ & $\mathbf{0.015}$ & $0.040$ \\
			\hline
		\end{tabular}
	\end{center}
	\vspace{-0.4cm}
\end{table}

\section{RIEMANNIAN MOTION TRANSFER}
\label{sec:MotionTransfer}
Understanding the underlying mechanisms of human motion generation is relevant not only to improve our overall comprehension of the human brain and body, but to generate efficient, well-coordinated, human-inspired robot motions. 
In general, transferring human movements to robots requires solving the correspondence problem~\cite{Nehaniv02:ImitationInAnimals}, i.e., finding a mapping between the different embodiments.
Such mappings were manually defined for similar kinematics~\cite{Mandery2016b:MotionDatabase} or learned automatically from data~\cite{aberman_skeleton-aware_2020}.
To bypass the complexity of these mappings, several works instead focused on transferring the functional part of the motion by mapping end-effector trajectories~\cite{Rakita17:MotionTransfer} or manipulability patterns~\cite{Jaquier20:ManipAnalysis}.

\begin{figure}
    \begin{subfigure}[b]{0.26\textwidth}
    \centering
    	\includegraphics[width=\textwidth]{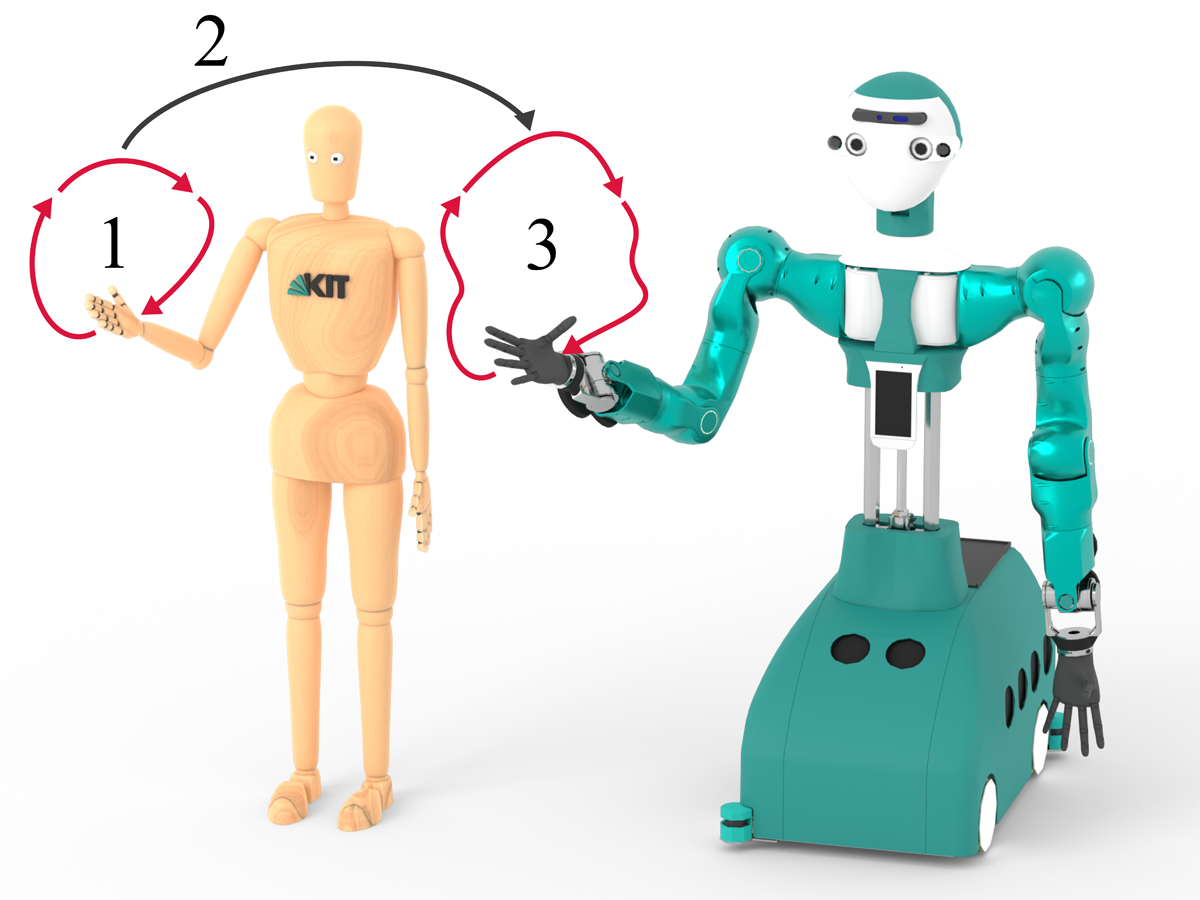}
    	\caption{Proposed transfer framework}
    	\label{Fig:TransferIllustration}
    \end{subfigure}
    \begin{subfigure}[b]{0.22\textwidth}
    \centering
	\includegraphics[width=0.95\textwidth]{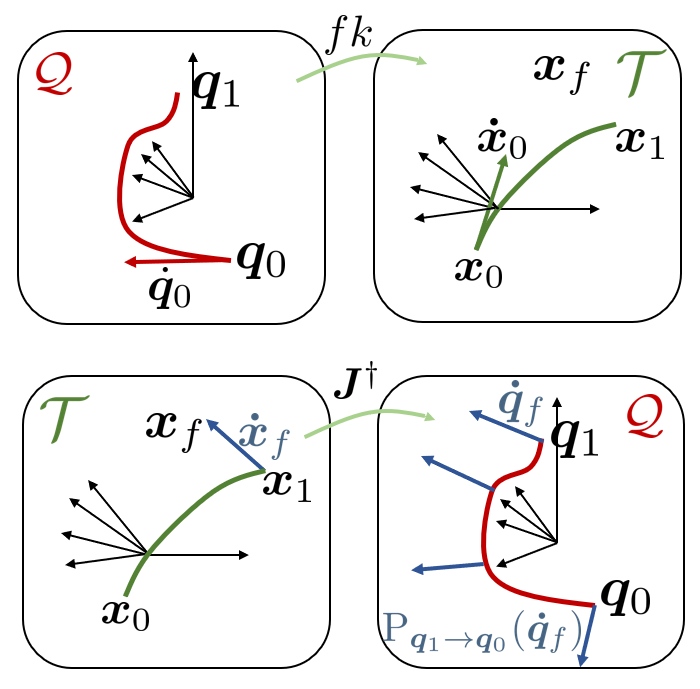}
	\caption{Transfer of geodesic synergy}
	\label{Fig:TaskspaceExpMapGradient}
	\end{subfigure}
	\caption{\emph{(a)} Illustration of the proposed transfer framework. (1) A human motion is segmented onto geodesic synergies. (2) The start and end points of each synergy are transferred onto the robot task space. Small synergies are merged together. (3) Each synergy is reproduced as a geodesic in the robot configuration manifold. \\
	\emph{(b)} Illustration of the transfer of a geodesic synergy from the human to the robot configuration space. \emph{Top:} 
	A geodesic in the human configuration space leads to a trajectory in task space going from $\bm{x}_0$ to $\bm{x}_1$.
	\emph{Bottom-left:} The direction from the current pose $\bm{x}_1$ to the desired one $\bm{x}_f$ is computed in task space and pulled back onto $\jointvelocity_f$ in the robot joint space. \emph{Bottom-right:} The gradient in joint space is then obtained through the parallel transport of $\jointvelocity_f$.}
	\vspace{-0.5cm}
\end{figure}

In this section, we propose to further exploit our Riemannian computational model to similarly transfer human movements to humanoid robots without the need of complex kinematic mappings.
We assume that the key points of a motion are given by the start and end of the underlying geodesic synergies. Therefore, we use the Riemannian segmentation introduced in Section~\ref{subsec:MotionSegmentation} to extract geodesic synergies from an observed human motion. Each extracted human synergy is then reproduced as a geodesic in the robot configuration manifold, as explained shortly. 
Therefore, each transferred synergy constitutes a minimum-energy trajectory accounting for the robot's own inertial properties~\cite{Sekimoto08:RiemannianDistanceBasedMotionPlanning}, while the overall sequence of geodesics conserves the main characteristics of the original human motion.
The proposed transfer framework is illustrated in Fig.~\ref{Fig:TransferIllustration} and evaluated for one human arm motion analyzed in Section~\ref{sec:MotionAnalysis}.

\subsection{Computation of geodesic synergies in the robot task space}
\label{subsec:RobotGeodesicSynergies}
Most purposeful human motions are generated to produce a desired hand trajectory. Therefore, when transferring an human motion to a robot, the key points of the trajectory must be replicated in its task space $\mathcal{T}$.
Given the Riemannian segmentation of a human motion, this corresponds to finding geodesic synergies in the robot configuration manifold, so that they connect the start and end of the human synergies in task space. Here, we consider the hand positions and orientations, i.e., $\mathcal{T} \sim \mathbb{R}^3 \times \mathcal{S}^3$, with $\mathcal{S}^3$ the quaternion space.

Given the desired initial and final hand poses $\bm{x}_i^{(g)},\bm{x}_f^{(g)} \in\mathbb{R}^3 \times \mathcal{S}^3$ of the $g$-th geodesic synergy, we aim at finding a spatial path $\jointposition(s)$ in the robot configuration manifold satisfying $fk(\jointposition_0) = \bm{x}_i^{(g)}$ and $fk(\jointposition_1)=\bm{x}_f^{(g)}$, where $fk(\cdot)$ denotes the robot forward kinematics function.
As the geodesic synergies are executed one after an other, we can assume that the initial condition $fk(\jointposition_0) = \bm{x}_i^{(g)}$ is already satisfied (for $g=1$, an initial configuration may be obtained with, e.g., IK). 
Therefore, given an initial configuration $\jointposition_0$, we aim at finding the initial velocity $\jointvelocity_0$ which minimizes the error between the hand pose along the corresponding geodesic at $t=1$, i.e., $\bm{x}_1=fk(\bm{q}_1)$, and the desired hand pose $\bm{x}_f^{(g)}$.
This corresponds to solving the following optimization problem
\begin{equation}
    \min_{\jointvelocity_0\in\configtangentspace{\jointposition_0}} \| \logmap{\bm{x}_1}{\bm{x}_f} \|^2 \;\;\text{ with } \;\; \bm{x}_1 = \expmap{\jointposition_0}{\jointvelocity_0},
    \label{Eq:OptimizationRetargeting}
\end{equation}
where the Riemannian distance between $\bm{x}_1, \bm{x}_f^{(g)}\in\mathbb{R}^3 \times \mathcal{S}^3$ is equal to the norm of the logarithmic map. Although the problem~\eqref{Eq:OptimizationRetargeting} does not yield an analytical solution, it can be efficiently solved using an iterative gradient descent. Notice that a Euclidean algorithm can be used, as $\jointvelocity_0$ always belongs to the same tangent space $\configtangentspace{\jointposition_0}$. The gradient of~\eqref{Eq:OptimizationRetargeting} is computed using the chain rule by first calculating the derivative of the Riemannian distance as
\begin{equation}
    \frac{\partial \| \logmap{\bm{x}_1}{\bm{x}_f^{(g)}} \|^2}{\partial \bm{x}_1}
    = 2 \; \logmap{\bm{x}_1}{\bm{x}_f^{(g)}},
\end{equation}
then pulling back the desired change from task space to joint space by using the pseudo-inverse of the Jacobian $\bm{J}^\dagger$, and approximating the derivative of the exponential map using the parallel transport operation as shown in~\cite{Kim14:RiemannianMGLM}, so that
\begin{equation}
    \frac{\partial \|\logmap{\bm{x}_1}{\bm{x}_f^{(g)}}\|^2}{\partial \jointvelocity_0}
    \approx -\prltrsp{\jointposition_1}{\jointposition_0}{2 \; \bm{J}(\jointposition_1)^{\dagger} \logmap{\bm{x}_1}{\bm{x}_f^{(g)}}}.
\end{equation}
The corresponding procedure is illustrated in Fig.~\ref{Fig:TaskspaceExpMapGradient}.
After computing the spatial path of a geodesic synergy~\eqref{Eq:OptimizationRetargeting}, its temporal path is defined by the minimum energy time course~\eqref{Eq:GeodesicTemporalPath} along the obtained geodesic, as for human motions. 
Note that the boundary conditions of~\eqref{Eq:GeodesicTemporalPath} ensure the equality of the velocity magnitudes at the start of a synergy and at the end of the previous one, thus preventing sharp velocity changes.

\subsection{Results}
\begin{figure}[tbp]
	\centering
	\begin{subfigure}[b]{0.16\textwidth}
		\adjustbox{trim={.15\width} {.45\height} {0.2\width} {.1\height},clip}{\includegraphics[width=1.6\textwidth]{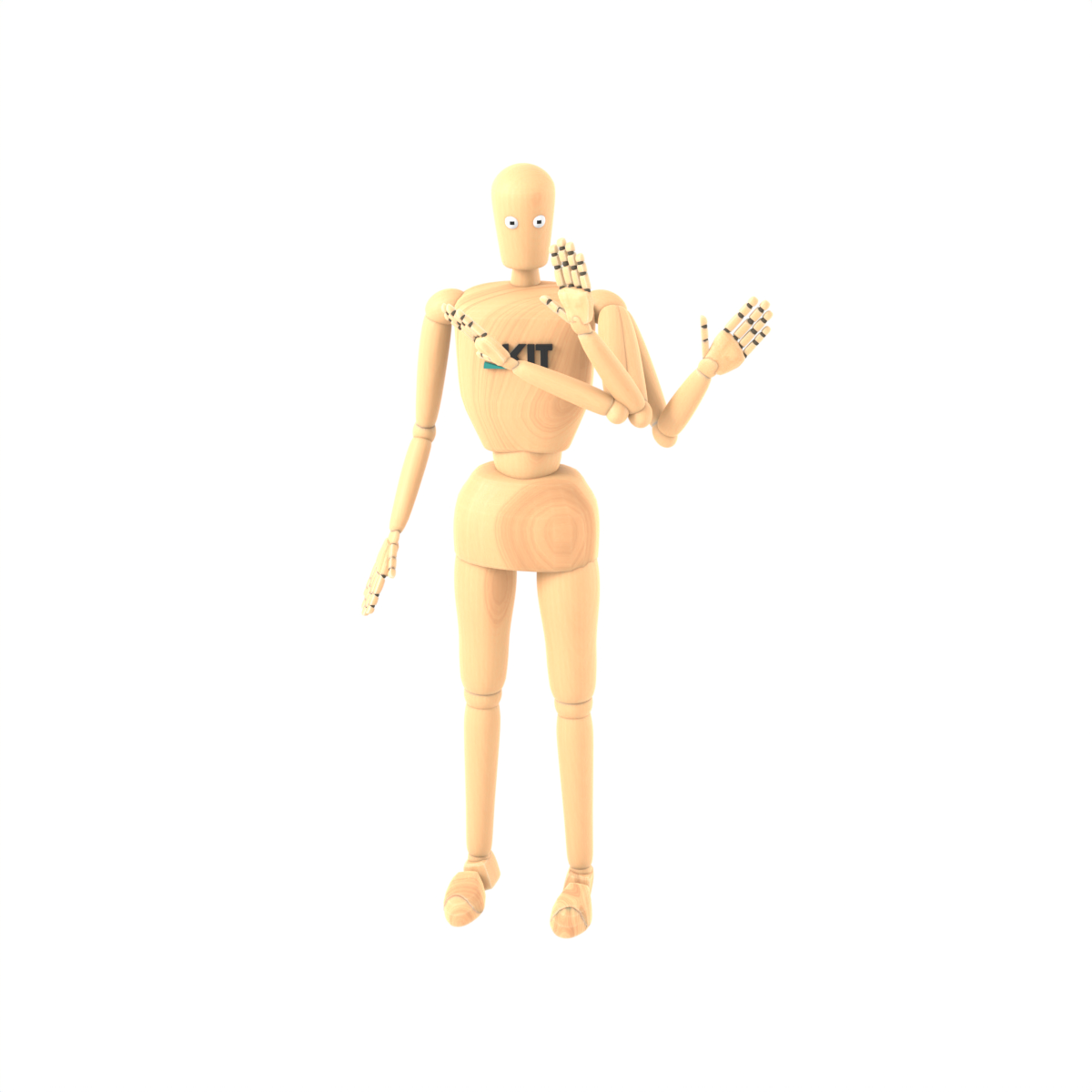}}
		\adjustbox{trim={.15\width} {.37\height} {0.2\width} {.12\height},clip}{\includegraphics[width=1.52\textwidth]{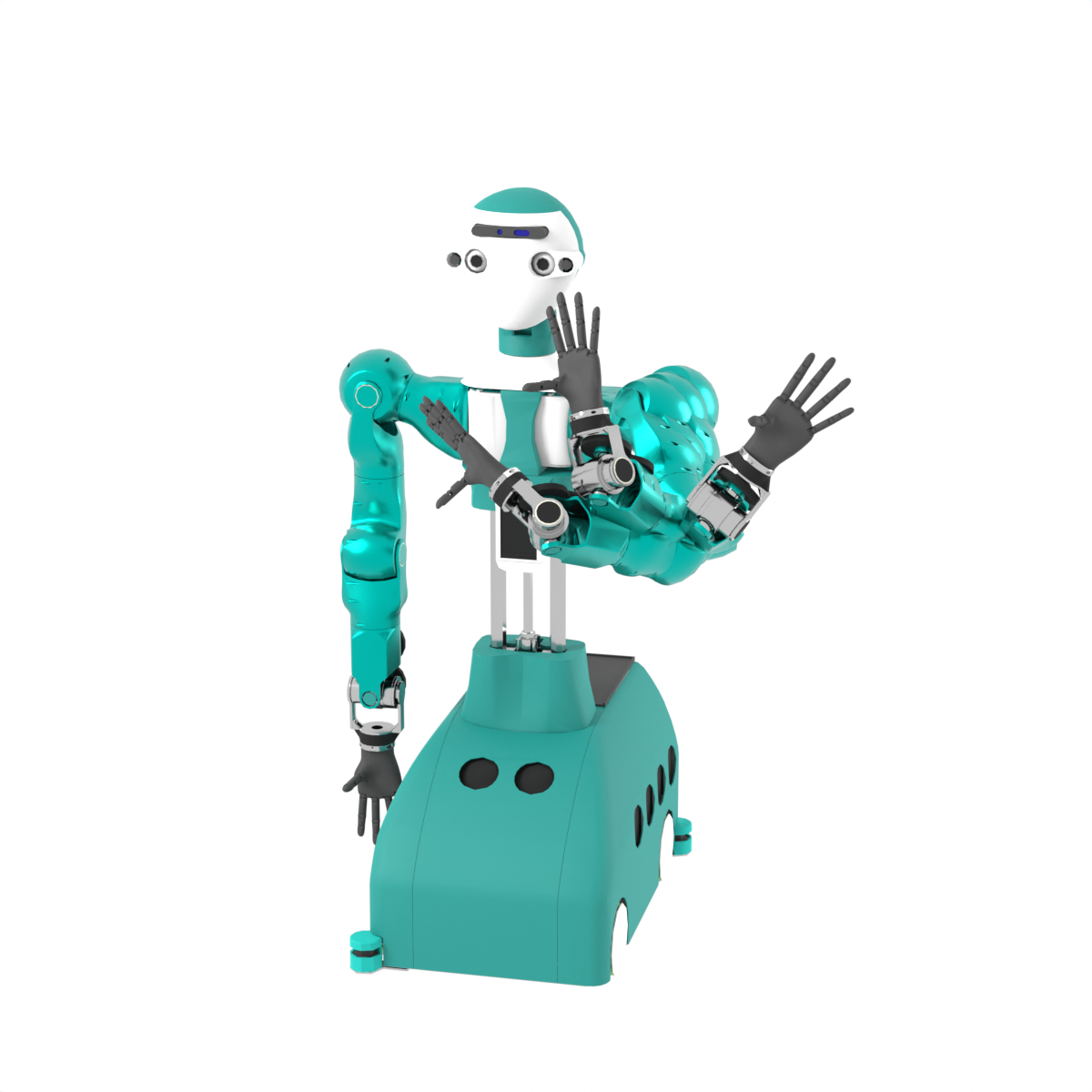}}
		\adjustbox{trim={.15\width} {.37\height} {0.2\width} {.12\height},clip}{\includegraphics[width=1.52\textwidth]{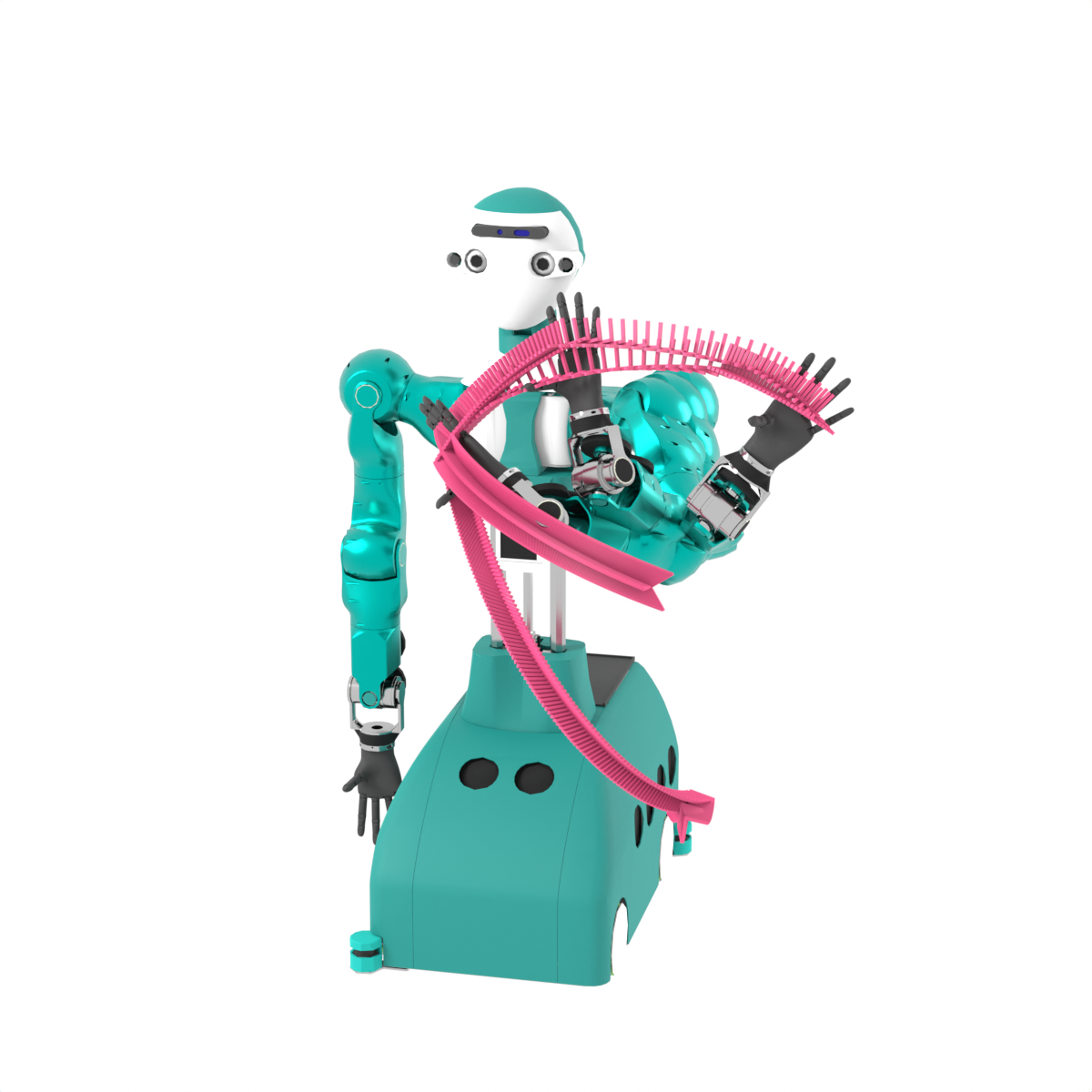}}
		\adjustbox{trim={.15\width} {.37\height} {0.2\width} {.12\height},clip}{\includegraphics[width=1.52\textwidth]{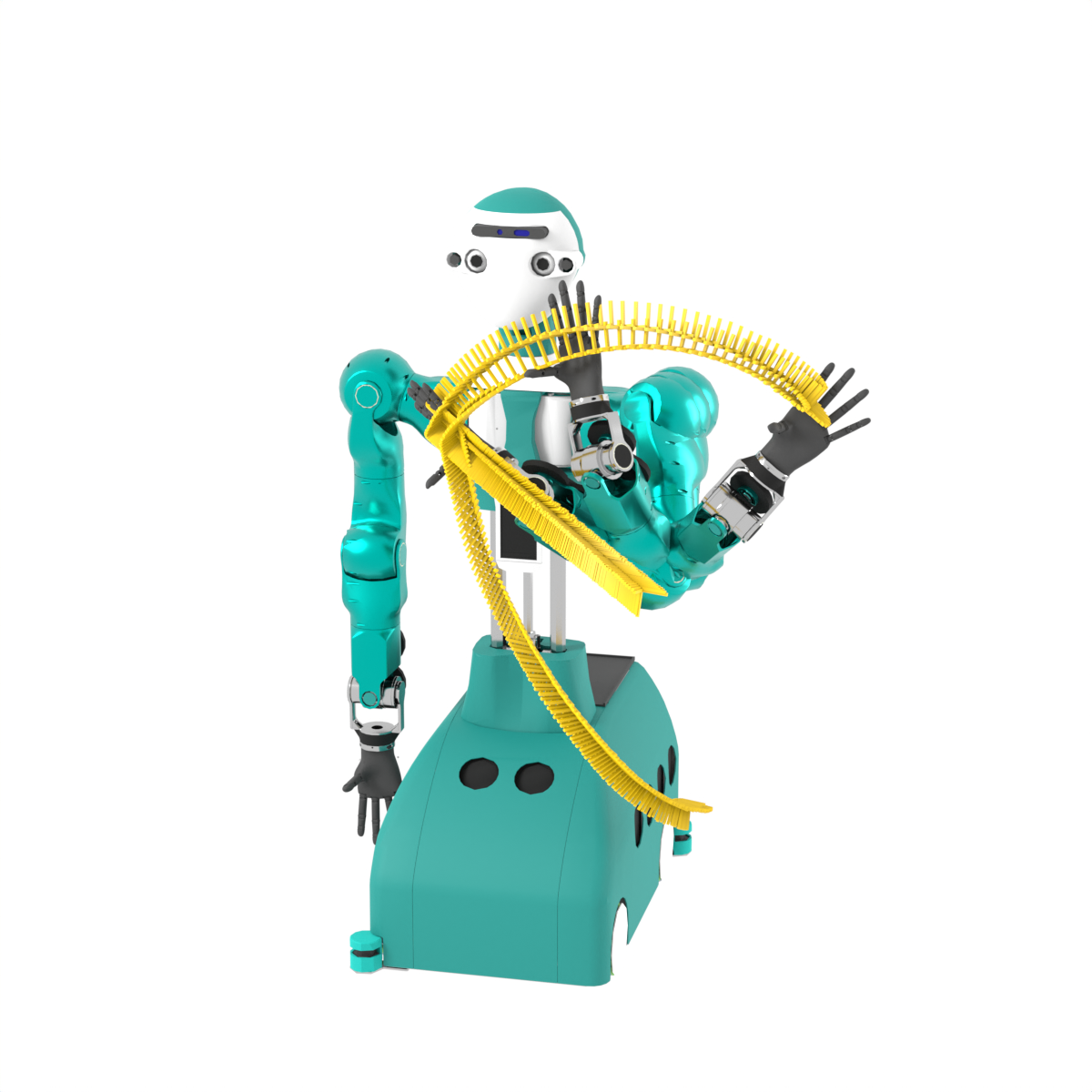}}
		\caption{$\mathsf{Waving}$ transfer}
		\label{subFig:TransferImages}
	\end{subfigure}
	\begin{subfigure}[b]{0.29\textwidth}
		\includegraphics[width=\textwidth,trim={0cm 0cm 0cm 0cm},clip]{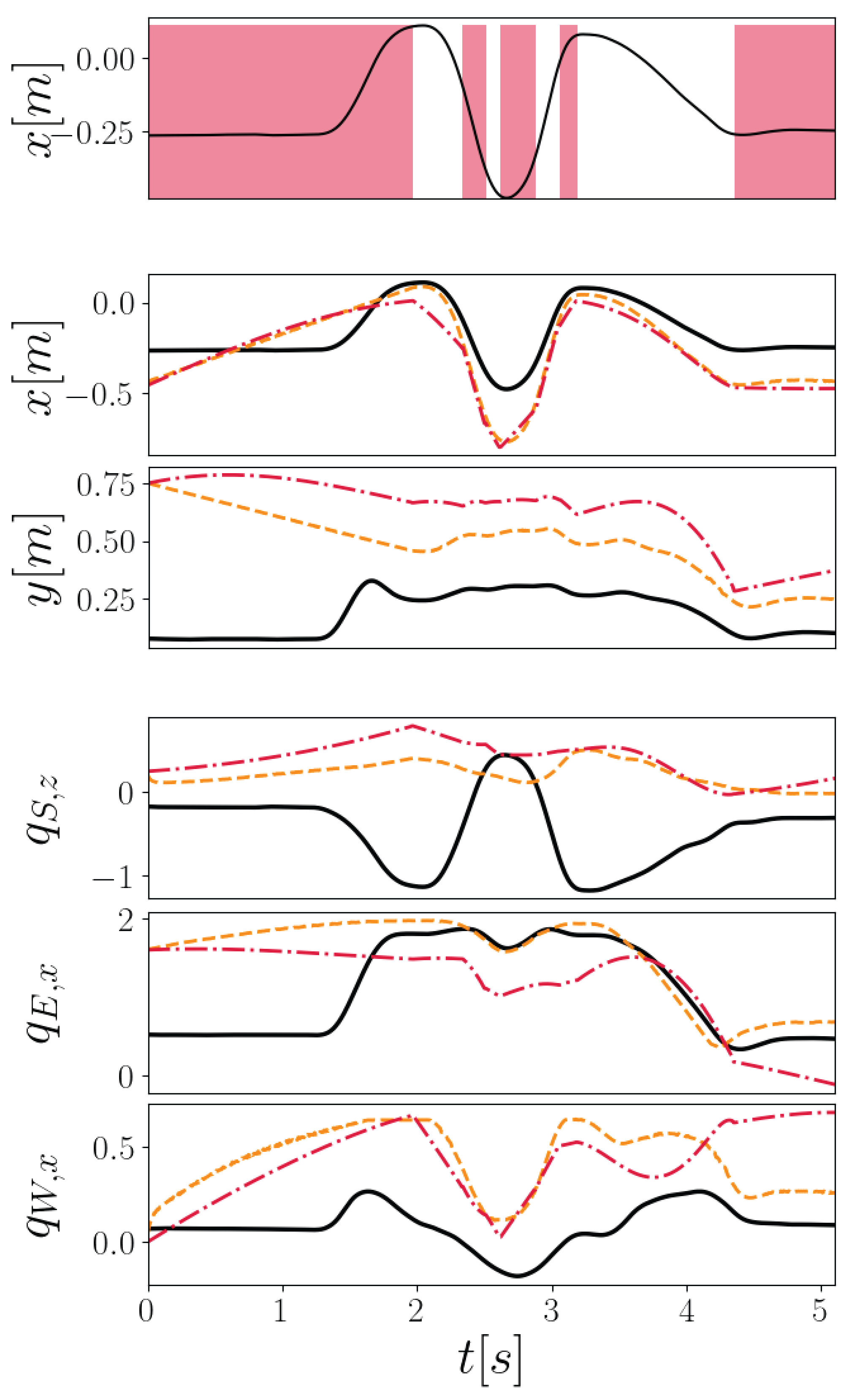}
		\caption{Segmentation and transferred motions}
		\label{subFig:TransferGraphs}
	\end{subfigure}
	\caption{Transfer of the short $\mathsf{waving}$ motion to \armarVI. \mbox{\emph{(a) - top-to-bottom}}: Ground-truth posture of the human, corresponding posture of the robot with the Riemannian model, hand trajectories obtained with the Riemannian, and IK model. The joint configurations are displayed at times $t=\{2, 2.35, 2.7\}$s. \emph{(b) - top row}: Merged Riemannian segmentation (\redrectangle \redrectangle) with the observed human hand trajectory (\blackline) along a relevant axis. Synergies are denoted by alternated colors. \emph{(b) - 2 middle rows}: Human (\blackline) and robot hand paths transferred with the Riemannian (\redline), and IK (\orangeline) models along two axes. \emph{(b) - 3 bottom rows}: Human and transferred trajectories for the shoulder, elbow and wrist joints $q_{S,z}$, $q_{E,x}$, $q_{W,x}$.}
	\label{Fig:TransferResults}
	\vspace{-0.4cm}
\end{figure}

Here we evaluate the proposed framework by transferring the short $\mathsf{waving}$ motion analyzed in Section~\ref{sec:MotionAnalysis} to the 8-DoFs arm of the humanoid robot \armarVI~\cite{Asfour2019:armar6}. In order to account for the difference of length between the human and robot arms, the hand poses are encoded in the shoulder frame and scaled in function of the total arm length. Moreover, we discard very small synergies and merge small ones together, so that they cover a distance higher than a given threshold in joint space. As shown in Fig.~\ref{subFig:TransferGraphs}-\emph{top}, this process conserves the long synergies encoding the core parts of the motions, while simplifying the transitions.

The joint configurations obtained for a shoulder, an elbow, and a wrist joint when transferring the $\mathsf{waving}$ motion to \armarVI are depicted in Fig.~\ref{subFig:TransferGraphs} (3 bottom rows) and visualized in Fig.~\ref{subFig:TransferImages}. The corresponding hand trajectories are displayed for the most relevant dimensions in the 2 middle rows of Fig.~\ref{subFig:TransferGraphs}. The joint configurations and hand trajectories obtained with IK are depicted as a baseline.

We observe that the main features of the $\mathsf{waving}$ motion are conserved by both the Riemannian transfer and IK. Although the obtained hand poses look similar, the underlying joint configurations slightly differ due to the fact that the Riemannian transfer follows piecewise minimum-energy trajectories along the motion. As expected the human and obtained robot joint configurations differ. This can be explained by the different kinematic and dynamic properties of the two agents.

\section{DISCUSSION}
\label{sec:Conclusion}
This paper presented a detailed analysis of human arm motions under the hypothesis that human movements are planned as sequences of geodesic synergies. 
Importantly, our analysis showed that the core parts of human arm motions (e.g., lifting, putting down the hand, or waving left-to-right) can be efficiently represented by one or two geodesic synergies. In contrast, transitions, including the start and end of the motions, are usually encoded by several short synergies. This can be explained by the various changes of trajectories observed during the transitions, which contrast with the smooth and fast movements characterizing the core parts of human motions.
Additionally, the proposed Riemannian computational model successfully predicted the spatial and temporal characteristics of the joint and hand trajectories of observed human motions. In contrast, the equivalent Euclidean model, generating minimum-energy trajectories in Euclidean space, induced various prediction errors. 
Therefore, our study confirms the high influence of the dynamical properties of the human body for motion generation. It further extends the results of~\cite{Biess07:ComputationalModelPointing,Sekimoto09:RiemannianDistanceBasedHumanMotion,Biess11:MovementDynamicsOptimizationInvariance} beyond point-to-point motions and validates that complex everyday human arm motions may be planned as sequences of minimum-energy movements in the configuration manifold.

It is important to note that our analysis is based on the normalized MMM reference model~\cite{Mandery2016b:MotionDatabase}, whose proportions and inertias are based on average measurements. Although these properties are scaled in function of the height and weight of each subject, they do not account for other subject-specific measurements. Therefore, the mass-inertia matrix of the MMM model may slightly differ from the one of the humans who executed the recorded motions. Additionally, small differences between the joint configurations of the human and the MMM model may arise due to the mapping from the motion capture system to the MMM model. 
This may result in slight differences in the Riemannian segmentation and in the computation of geodesic synergies.

With the presented Riemannian transfer framework, this paper introduced a novel perspective on the motion retargeting problem. Namely, our transfer framework naturally generate energy-optimal motions with respect to the dynamical properties of the considered agents. %
In this paper, we exploited key points in task space to determine the geodesic synergies for transferred motions. In our future work, we will explore if the mapping of synergies from the human to the robot configuration manifold can instead be learned for specific tasks. Moreover, the proposed transfer framework does not incorporate joint limits and self-collisions avoidance. This may be achieved, e.g., by artificially modifying the Riemannian metric around problematic joint configurations. 

In this study, the presented analysis and transfer frameworks were restricted to the 7 DoFs of the human arm. As the considered motions mostly involve the movement of a single arm, the influence of the remaining DoFs of the body on the planned geodesic synergies remains limited. However, many human motions involve the coordination of multiple joints throughout the whole body, e.g., for bimanual manipulation or loco-manipulation tasks~\cite{Krebs21:MotionDatabase}. As discussed by Neilson et al.~\cite{Neilson15:GeodesicSynergyHypothesis}, whole-body motions may not be solely explained by single geodesic synergies, but instead by several synergies activated \emph{simultaneously}. Thus, we may hypothesize that whole-body motions are planned as sequences of combined geodesic synergies. Future work will therefore generalize the Riemannian framework presented in this paper to analyze and transfer whole-body human motions including combinations of geodesic synergies. 


\bibliographystyle{IEEEtran}

\bibliography{References} 

\end{document}